\documentclass{article}

\usepackage[accepted]{icml2026}

\usepackage[utf8]{inputenc} % allow utf-8 input
\usepackage[T1]{fontenc}    % use 8-bit T1 fonts
\usepackage{hyperref}       % hyperlinks
\usepackage{url}            % simple URL typesetting
\usepackage{booktabs}       % professional-quality tables
\usepackage{amsfonts}       % blackboard math symbols
\usepackage{amsthm}        % for theorem and proof environments
\usepackage{nicefrac}       % compact symbols for 1/2, etc.
\usepackage{microtype}      % microtypography
\usepackage{xcolor}         % colors
\usepackage{subcaption}
\usepackage{tabularx} 
\usepackage{graphicx}
\usepackage{wrapfig,lipsum,booktabs}
\usepackage{amsmath}
\usepackage{amssymb}
\usepackage{mathtools}
\usepackage{etoc}
\usepackage{comment}
\usepackage{paralist,xcolor}

\usepackage{tikz}
\usetikzlibrary{shapes, arrows.meta, positioning, calc, fit}

\usepackage[export]{adjustbox}
\usepackage{multirow}
\theoremstyle{plain}

\theoremstyle{definition}

\theoremstyle{remark}

\newcommand{\Col}[2]{{\color{#1}#2}}
\newcommand{\Note}[2]{[\textbf{#1}: #2]}
\newcommand{\elenic}[1]{\Col{purple}{#1}}
\newcommand{\eleni}[1]{\elenic{\Note{Eleni}{#1}}}

\newcommand{\peterc}[1]{\Col{orange}{#1}}
\newcommand{\peter}[1]{\peterc{\Note{Peter}{#1}}}
\newcommand{\ci}[2]{#1\,\ensuremath{{\scriptstyle \pm #2}}}

\begin{document}
\twocolumn[
  \icmltitle{ You Don't Need All That Attention: \\
  Surgical Memorization Mitigation in Text-to-Image Diffusion Models}
\icmltitlerunning{Surgical Memorization Mitigation in Text-to-Image Diffusion Models}



  \icmlsetsymbol{equal}{*}

  \begin{icmlauthorlist}
    \icmlauthor{Kairan Zhao}{yyy}
    \icmlauthor{Eleni Triantafillou}{comp}
    \icmlauthor{Peter Triantafillou}{yyy}
    % \icmlauthor{Firstname1 Lastname1}{equal,yyy}
    % \icmlauthor{Firstname2 Lastname2}{equal,yyy,comp}
    % \icmlauthor{Firstname3 Lastname3}{comp}
  \end{icmlauthorlist}

  \icmlaffiliation{yyy}{University of Warwick, UK}
  \icmlaffiliation{comp}{Google DeepMind, UK}
  % \icmlaffiliation{equal}{Equal senior contribution}

  \icmlcorrespondingauthor{Kairan Zhao}{Kairan.Zhao@warwick.ac.uk}

  \icmlkeywords{Machine Unlearning, ICML}

  \vskip 0.3in
]

\printAffiliationsAndNotice{}  % no special notice (required even if empty)

% \maketitle

\begin{abstract}
Generative models have been shown to ``memorize'' certain training data, leading to 
verbatim or near-verbatim 
generating images, which may cause privacy concerns or copyright infringement. We introduce Guidance Using Attractive-Repulsive Dynamics (GUARD), a novel framework for memorization mitigation in text-to-image diffusion models. GUARD adjusts the image denoising process to guide the generation away from an original training image and towards one that is distinct from training data while remaining  aligned with the prompt, guarding against reproducing training data, without hurting image generation quality. 
We propose a concrete instantiation of this framework, where the positive target that we steer towards is given by a novel method for  (cross) attention attenuation based on (i) a novel statistical mechanism that automatically identifies the prompt positions where cross attention must be attenuated and (ii) attenuating cross-attention in these per-prompt locations. 
The resulting GUARD offers a surgical, dynamic per-prompt inference-time approach that, we find, is by far the most robust method in terms of consistently producing state-of-the-art results for memorization mitigation across two architectures and for both verbatim and template memorization, while also improving upon or yielding comparable results in terms of image quality.
% can significantly outperform all prior work in terms of memorization mitigation and image-generation quality preservation, across two architectures (SD v1.4 and SD v.2.0) and for both verbatim and template memorization.
\end{abstract}

\section{Introduction}

Researchers have recently observed that generative models can reproduce verbatim or near-verbatim copies of certain training examples \citep{carlini2021extracting,carlini2023extracting,somepalli2023diffusion}; a phenomenon referred to as ``memorization''. This is problematic for various reasons, ranging from revealing sensitive or private data to potential copyright concerns, and numerous research efforts have attempted to address the issue of memorization either preventatively or post-hoc. We study this important problem in the context of text-to-image (T2I) diffusion models.  

Several approaches have been proposed to address memorization, including ``training time'' methods that act during the original training phase, to prevent memorization from occurring in the first place \citep{ren2024unveiling,wen2024detecting}, and methods operating at post hoc, ``finetuning time''  that aim to post-process a model to remove the memorization of certain examples (``unlearning'') \citep{golatkar2020eternal,kurmanji23,l1sparse,salun,zhao2024,alberti2025data}. However, these methods have various drawbacks: in modern pipelines, practitioners often build on pretrained models and lack control of the original pretraining phase, making training-time interventions unrealistic. Even setting that unrealistic assumption aside, training-time interventions that can at best guess which examples will become memorized, rely on heuristics and may have unwanted side-effects such as deteriorating the model's utility. Finetuning-based unlearning methods, on the other hand, are both computationally inefficient and often lack robustness \citep{lucki2024adversarial,siddiqui2025dormant}.

Given the above shortcomings of training- and finetuning- time methods, and the desiderata of computational efficiency and surgical precision of memorization reduction without compromising utility, we turn our attention to a third category of inference-time mitigation: rather than seeking a model that has no memorized information in its weights, we accept that memorization may have occurred, but devise specialized inference-time procedures that mask memorized information so that it doesn't affect the generated content. 
In other words, while some memorized information may be present in the model weights, we seek a specialized forward pass at inference-time that: 
(i) guards against generating any memorized training example; 
(ii) does so in a manner that does not adversely affect the high-quality of generated images; and 
(iii) with minimal impact to efficiency.  

Our first contribution is a novel inference-time mitigation framework for T2I diffusion models called Guidance Using Attractive-Repulsive Dynamics (GUARD). 
% a ``contrastive guidance'' mechanism for the denoising procedure of the diffusion model. 
We build on the classic ``classifier-free guidance'' formula, that sets the predicted noise for a given timestep of denoising to be a combination of the predicted noise obtained from an empty prompt (the ``unconditional prediction'') and the predicted noise obtained from the given text prompt (the ``text-conditional prediction'', obtained via cross-attention with the text prompt embeddings). We modify this guidance to add a \textit{negative} weight to the text-conditional  prediction (the ``repulsion'' term), and we add an additional ``attraction'' term with a positive weight. Effectively performing arithmetic 
% in the latent space 
of the predicted noise, our revised ``contrastive guidance'' is governed by two forces. The ``repulsive'' term encourages moving away from the predicted noise that the standard forward pass would have produced on the given prompt. This is a key driver of memorization mitigation, since that standard forward pass applied on a memorized prompt would steer the generation towards  reconstructing the training image associated with that prompt. At the same time, the ``attraction term'' provides a better target to steer towards. This term not only helps with mitigating memorization but acts as a quality booster, as steering aggressively away from the memorized prompt can otherwise  cause fidelity collapse, where generated images lose structural coherence or semantic relevance to the prompt. 

Next, we propose a concrete instantiation of the GUARD framework via a novel cross-attention adjustment method playing the role of the ``positive target'' that GUARD steers towards. We arrived at this method after a careful empirical analysis of the distribution of cross-attention at particular tokens, building and expanding on past work that found that specific ``trigger'' tokens are responsible for regurgitating training images \cite{somepalli2023diffusion, wen2024detecting}. Compared to prior work that uses hard-coded strategies to redistribute the (cross)-attention \citep{ren2024unveiling}, we develop a dynamic on-the-fly and per-prompt trigger-token detection strategy to inform attention redistribution. 
We demonstrate that this approach improves upon the prior state of the art on its own, but even by a larger margin when integrated into GUARD as the positive target.

We summarize our key contributions below. 
% \vskip -0.2in
\begin{itemize}
    \item \textbf{Framework:} We propose GUARD, a contrastive guidance framework for inference-time memorization mitigation that combines repulsion from memorized  directions with attraction toward a safe target.
    \item \textbf{Empirical Analysis.} We analyze the distribution of cross attention at tokens of memorized and non-memorized prompts, across model architectures and for verbatim and template memorization, revealing more nuanced desiderata for memorization mitigation.
    \item \textbf{Detection:} Guided by our analysis, we design a per-prompt cross-attention spike detector that identifies memorization-critical positions $\mathcal{S}(p)$ for a prompt $p$ based on on-the-fly statistical outlier analysis.
    \item \textbf{Instantiation:} We instantiate GUARD with a surgical cross-attention-logit attenuation mechanism that attenuates attention at positions in $\mathcal{S}(p)$. 
    \item \textbf{Evaluation:} We conduct a comprehensive evaluation of GUARD and its instantiations versus several prior state-of-the-art methods.  The results show that our method outperforms prior work across model architecture, memorization types (verbatim vs template) and according to both memorization and quality metrics.

\end{itemize}

\section{Background} \label{sec:background}

\textbf{Diffusion Models}. 
In image diffusion models, a ``forward diffusion'' process first adds Gaussian noise to an image. This happens gradually over $T$ steps, with each step adding a predetermined amount of noise. The image $x_t$ at step $t$ of this forward process can be computed in closed form as: 
% \vskip -0.1in
\begin{equation}
% x_t = \sqrt{\alpha_t} x_0 + \sqrt{1 - \alpha_t} \epsilon
x_t = \sqrt{\bar\alpha_t} x_0 + \sqrt{1-\bar\alpha_t} \epsilon,
\end{equation}
% \vskip -0.1in
where $x_0$ is the original image (before any noise addition), and $\bar\alpha_t = \prod_{i=1}^t (1 - \beta_t)$, where $\beta_t \in (0,1)$ is the scheduled variance at step $t$, and $\epsilon \sim \mathcal{N}(0,I)$.

Then, the ``reverse diffusion'' process gradually removes the noise, aiming to obtain the original clean image. This happens via a trainable ``noise predictor'' model $\epsilon_\theta$, parameterized by $\theta$, that takes as input a noised image and predicts the noise that was added to it. 
% In closed form, we can compute: 
% \begin{equation}
% x_{t-1} = \sqrt{\alpha_{t-1}} x_0 + \sqrt{1 - \alpha_t} \epsilon
% \end{equation}
The training objective for parameters $\theta$ is then the MSE loss between the sampled noise and the predicted noise across timesteps $t$:
% \vskip -0.1in
\begin{equation} \label{eq:standard_loss}
\mathcal{L}(\theta) = \mathbb{E}_{t,\,x_0,\,\epsilon \sim \mathcal{N}(0,1)}\;\big\|\epsilon_{t} - \epsilon_{\theta}(x_t,t)\big\|^2_2
\end{equation}
% \vskip -0.1in

\textbf{Text-to-Image Diffusion Models}. 
Some text-conditional diffusion models, such as Stable Diffusion, use classifier-free guidance \cite{rombach2022high} at sampling time to steer generations. During inference, given a text prompt \(p\), it 
% In this case, the above process is conditioned on a text prompt $p$, that 
is first embedded into a latent space using a text encoder $f$, to obtain the prompt embedding $e_p = f(p)$. The predicted noise at step \(t\) is now
\begin{equation} \label{eq:text_cond_pred}
\epsilon_{\theta}(x_t,t,e_p) = \epsilon_{\theta}\bigl(x_t, t, e_{\emptyset}\bigr)
\;+\;
s\bigl(\epsilon_{\theta}(x_t, t, e_p)
-\epsilon_{\theta}(x_t, t, e_\emptyset)\bigr)
\end{equation}
% \begin{equation} \label{eq:standard_loss}
% \mathcal{L} = \epsilon_{\theta}\bigl(x_t, e_{\emptyset}\bigr)
% \;+\;
% s\bigl(\epsilon_{\theta}(x_t, e_p)
% -\epsilon_{\theta}(x_t, e_\emptyset)\bigr)
% \end{equation}
where $e_\emptyset$ is the embedding of the empty prompt and $s$ represents the strength of the ``text guidance'', controlling how strongly the prediction should be influenced by the prompt.

In these models, typically the denoising network is a U-Net. 
A U-Net predicts a noise $\epsilon_\theta(x_t, t, e_p)$ from a noisy latent $x_t$ at timestep $t$, conditioned on the prompt embedding $e_p$. The U-Net consists of a ``downsampling'' path, a ``bottleneck'' (mid) block, and an ``upsampling'' path, with skip connections between corresponding resolutions. 
Text conditioning is achieved via cross-attention layers placed at multiple U-Net blocks, enabling spatial latent features to selectively attend to text-token embeddings.

Concretely, let $H \in \mathbb{R}^{N \times d}$ denote the sequence of latent features at some U-Net block (e.g., $N$ spatial positions flattened) and let
$E \in \mathbb{R}^{L \times d}$ denote the sequence of $L$ text token embeddings from the text encoder.
A standard cross-attention layer forms queries $Q = HW_Q \in \mathbb{R}^{N \times d_k}$, keys $K = EW_K \in \mathbb{R}^{L \times d_k}$, and values
$V = EW_V \in \mathbb{R}^{L \times d_v}$.
Attention logits are
$\ell = QK^\top/\sqrt{d_k} \in \mathbb{R}^{N \times L}$, and attention weights are
$A = \mathrm{softmax}(\ell)$ over the token, producing attended features $AV$.
In later sections we manipulate a subset of these logits $\ell_{q,i}$ for selected token indices $i$ (and queries $q$) prior to the softmax. 
This serves as a mechanism to reduce the weight of memorization-critical tokens, 
while leaving the rest intact.

\textbf{Memorization types}. Prior work identified and distinguished between two types of ``memorization'' in text-to-image diffusion models: ``verbatim memorization'' and ``template memorization'' \citep{webster2023reproducible,ren2024unveiling}. Verbatim memorization refers to the model reproducing near-identical copies of specific training instances when prompted with the original text. 
In contrast, template memorization manifests as the model reproducing images closely aligned with the training instances while allowing non-semantic variations. 
Although both phenomena reflect memorization, they arise from different mechanisms and exhibit distinct behavioral signatures. 
This is an important distinction as we will show later that prior methods designed for one type of memorization fail to mitigate the other type. We are after a general mitigation method that works across the board, so we design  with this desideratum in mind.%, and we conduct extensive evaluations on different memorization types in later sections.
% Let us first analyze the qualitative differences between verbatim and template memorization, which uncovers distinct generation patterns that help explain their differing mitigation difficulty.

% \vskip -0.2in
\section{Related Work}

% Addressing the detrimental effects of memorization has received a great deal of attention in the ML literature as a whole. 
Related efforts can be categorized into three broad categories; we present an overview below. 
% (i) \textit{Training-Time} Memorization Mitigation (TTMM) methods, (ii) \textit{Post-training finetuning-based} mitigation, under the name of ``machine unlearning'' and (iii) Inference-Time Memorization Mitigation (ITMM) methods.
% Methods in the latter category, in contrast to the former two, do not alter model weights; instead, they aim to, on-the-fly, avoid generating memorized training content. 
% Our work falls in inference-time mitigation, which has been attracting significant attention recently due to its promise to avoid generating memorized content on-the-fly in a computationally efficient manner, without otherwise deteriorating the model's capabilities. 
% In addition, related work can be classified according to whether they operate at the granularity of training examples or of more general concepts/entities. 

\textbf{Training Time Memorization Mitigation (TTMM)} methods intervene during training to prevent memorization from occurring \citep{ren2024unveiling,wen2024detecting}. If successful, no memorized information will be in the weights, making the resulting model robust against white-box extraction attacks. However, these methods are often ``blunt instruments'' that operate at a coarse granularity: because they operate during training when it's not clear which examples will become memorized, they may overly suppress example influence with detrimental consequences for learning or unintended side effects such as a utility degradation and the indiscriminate suppression of benign but rare visual concepts. 
Furthermore, in modern pipelines, it is common practice to leverage pretrained models whose training process we have no control over, making TTMM unrealistic.
% In this realm, while TTMM methods can offer protection against white-box attacks by editing model weights, these methods are often "blunt instruments" that operate at a coarse granularity. This global suppression can lead to unintended side effects, such as a degradation of utility (model performance of remaining data) and the indiscriminate forgetting of benign but rare visual concepts.

\textbf{Machine unlearning} methods aim to remove the influence of specific training data from models post-training, therefore naturally lending themselves to memorization mitigation. These methods do not assume control over the original training phase, overcoming that limitation of TTMM. Foundational unlearning algorithms were designed for image classification using techniques like NegGrad \citep{golatkar2020eternal}, random labeling \citep{graves2021amnesiac}, NegGrad+ \citep{kurmanji23},
SCRUB \citep{kurmanji23}, SalUn \citep{salun}, L1-Sparse \citep{l1sparse}, and RUM \cite{zhao2024} and later adapted to LLMs \citep{yao2024large,barbulescu24}. In the domain of text-to-image (T2I) diffusion models (DMs), most existing unlearning research focuses on concept-level forgetting, which aims to erase broad visual categories or artistic styles rather than individual examples \citep{gandikota2023erasing, cywinski25, chen25, schioppa24, zhang2024unlearncanvas, ko2024boosting, park2024direct}. Recently, \citet{alberti2025data} tackle the issue of example-level unlearning in text-to-image models, which is the most closely related to our work out of the unlearning methods. However, unlearning approaches come with several issues: they are computationally inefficient as they require a finetuning phase for each given forget set, and recent work shows that believed-to-be unlearned information often surfaces spontaneously, raising concerns about the effectiveness of these methods \citep{siddiqui2025dormant,hu2024unlearning,deeb2024unlearning,lucki2024adversarial}.

\textbf{Inference Time Memorization Mitigation (ITMM).} Unlike the previous categories, ITMM methods do not modify the weights. Instead, they operate on the fly at inference time. ITMM is particularly attractive due to its expected computational efficiency and its ability to operate without access to the original training phase.  
% ITMM strategies in T2I diffusion models achieve this by editing the denoising trajectory during sampling. 
\citet{wen2024detecting} discovered that memorized prompts produce abnormally large text-conditional noise prediction magnitudes and proposed scaling down the conditional signal during sampling. Furthermore, \citet{ren2024unveiling} demonstrated that memorization is mechanistically linked to peaky Cross-Attention (CA) distributions on ``trigger tokens,'' specifically summary tokens such as the end-ot-text (EOT) and padding tokens. 
They propose an inference-time memorization mitigation technique that redistributes cross-attention among tokens, indirectly. Specifically, before softmax normalization the method  reduces cross-attention at the EOT and padding tokens (to $-\infty$) and boosts it at BOT (the beginning of text token). 
We contribute an analysis that explains why and when this method performs poorly. 
% Our analysis motivates and guides the development of a new method, which is based on a statistical cross-attention spike detector, to locate the relevant trigger tokens on-the-fly and per prompt.
% they proposed masking these summary tokens or boosting the beginning-of-text token (BOT) to redistribute CA. 
More recently, \citet{han2025adjusting} attributed memorization to an ``attraction basin'' in the sample-time space where classifier-free guidance (CFG) pulls trajectories toward memorized outputs; they proposed adjusting the initial noise sample to facilitate an earlier escape from this basin \citep{jain2025classifier}. 
% \kairan{
Similarly, \citet{jeon2025understanding} offer a complementary geometric perspective, linking memorization in diffusion models to sharp regions of the learned probability landscape and proposing a sharpness-aware adjustment of the initial noise for mitigation.
% }
Our work augments related work by contributing a novel suite of surgical approaches to inference-time mitigation and comprehensively study their performance. 
While the EOT token is a primary summary shortcut, memorization is also driven by instance-specific trigger tokens \textit{that do not follow consistent categorical patterns} (i.e. they cannot be identified solely through linguistic or semantic classifications—such as Part-of-Speech (POS) tags or named entity types, a fact also supported by the work in \citep{wen2024detecting}).
Therefore, we propose a statistical ``CA spike detector'' to locate these triggers on-the-fly, leading to dynamic (per-prompt) and surgical (intervening on required positions) cross-attention adjustment that improves on prior work. 

\begin{figure*}[t]
    \centering
    
    % --- Top Row ---
    \begin{subfigure}[b]{0.49\textwidth}
        \centering
        % Just set height. LaTeX will scale width automatically.
        % \includegraphics[height=2.5cm, width=\linewidth]
        \includegraphics[width=0.8\linewidth]{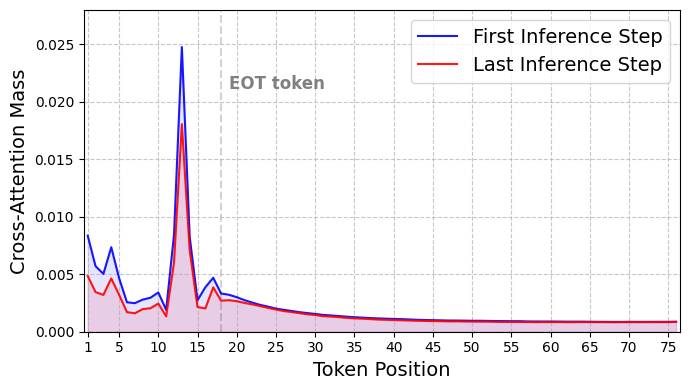}
        \caption{VM; CA distribution}% across tokens.}
        \label{fig:main-fig-distr-verbatim}
    \end{subfigure}
    % \hfill % Adds flexible space between the two images
    \begin{subfigure}[b]{0.49\textwidth}
        \centering
        % Just set height. LaTeX will scale width automatically.
        \includegraphics[width=0.8\linewidth]{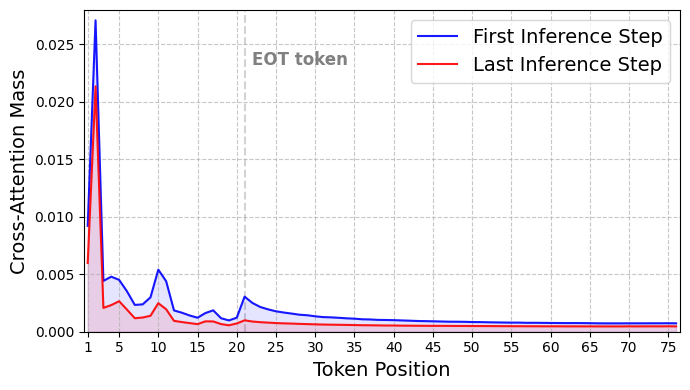}
        \caption{TM; CA distribution}% across tokens.}
        \label{fig:main-fig-distr-template}
    \end{subfigure}
    
   % --- Bottom Row ---
   
    \begin{subfigure}[b]{0.49\textwidth}
        \centering
        % Just set height. LaTeX will scale width automatically.
        % \includegraphics[height=2.5cm, width=\linewidth]
        \includegraphics[width=0.8\linewidth]{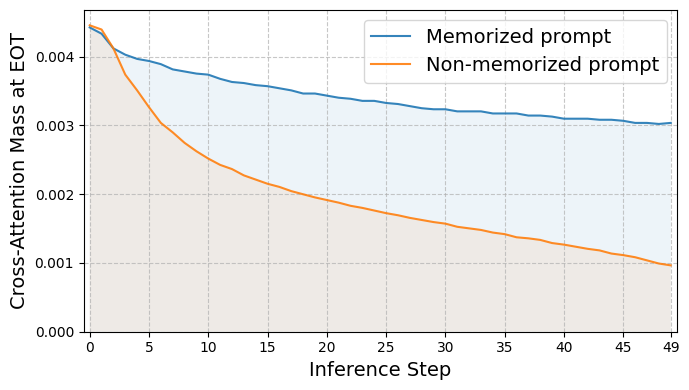}
        \caption{VM; CA mass on EOT}% across steps.}
        \label{fig:main-fig-ca-verbatim}
    \end{subfigure}
    % \hfill % Adds flexible space between the two images
    \begin{subfigure}[b]{0.49\textwidth}
        \centering
        % Just set height. LaTeX will scale width automatically.
        \includegraphics[width=0.8\linewidth]{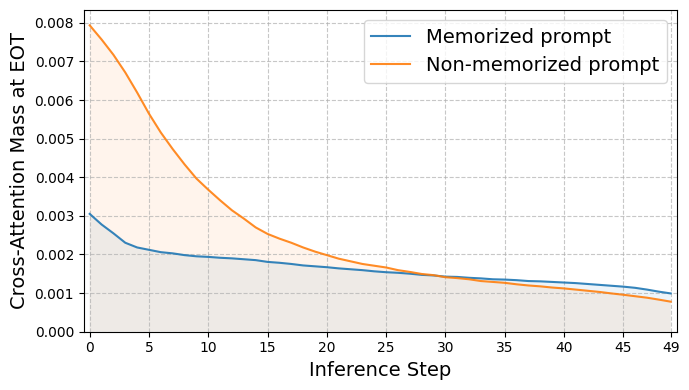}
        \caption{TM; CA mass on EOT}% across steps.}
        \label{fig:main-fig-ca-template}
    \end{subfigure}

    \caption{CA patterns. \textbf{(a) and (b)}: the CA mass across tokens, in the first and last inference steps for verbatim and template memorization (VM and TM). For clarity, we exclude the first token (position 0) from the plots, as it consistently receives the majority of CA across both memorized and non-memorized examples, which would dominate the scale and obscure differences among the remaining tokens. \textbf{(c) and (d)}: the CA mass on EOT across steps, for a memorized and a non-memorized prompt. \textbf{(c)}: VM on SD v1.4. \textbf{(d)}: TM on SD v2.0.
    }
    \label{fig:all-motivating-figs}
\end{figure*}

% \vskip -0.2in
\section{Introducing the GUARD}\label{sec:ar}
GUARD is an inference-time memorization mitigation framework that, given a text prompt, modifies the image denoising process in a specialized way to guard against reproducing an original training image, while still producing a high-quality and prompt-aligned image. 

GUARD achieves this via a modification of the standard 
% classifier-free guidance (
CFG. Specifically, it steers the generation away from memorized content (``repulsion'') and simultaneously, towards a high-quality but distinct-from-training-data alternative (``attraction'').
The attraction towards a positive target serves two purposes: (i) to lower memorization, due to providing a safe target to redirect towards, and (ii) to maintain high quality; without an alternative high-quality target to redirect to, the use of the repulsion term can otherwise inadvertently harm quality while reducing memorization. 

Let $\epsilon_\theta(x_t, e_{\phi})$ denote the unconditional noise prediction at denoising timestep $t$, and let $\epsilon_\theta^{-}(x_t, e_p)$ denote the noise prediction conditioned on the original (memorized) prompt.
Standard CFG computes the guided noise prediction as
% \vskip -0.1in
\begin{equation}
\hat{\epsilon} = \epsilon_\theta(x_t, e_{\phi}) + s \left( \epsilon_\theta^{-}(x_t, e_{p}) - \epsilon_\theta(x_t, e_{\phi}) \right)
\end{equation}
% \vskip -0.1in
where $s$ is the guidance scale. 
This allows to control the degree of adherence to the given prompt, 
% which is useful for producing images that are semantically relevant to the prompt, 
but may cause generating original training data. 
In contrast, GUARD redirects the generation away from the original prompt and towards a newly-added positive conditional noise prediction term $\epsilon_\theta^{+}(x_t, e_p)$, as follows:
% The goal is to generate content that remains semantically aligned with the prompt but without reproducing training data.
% The noise prediction of GUARD is then given by
% \vskip -0.1in
\begin{equation}
\begin{aligned}
\hat{\epsilon}
= {} & \epsilon_\theta(x_t, e_{\phi})
+ s \big( \epsilon_\theta^{+}(x_t, e_p) - \epsilon_\theta(x_t, e_{\phi}) \big) \\
& - r \big( \epsilon_\theta^{-}(x_t,e_p) - \epsilon_\theta(x_t, e_{\phi}) \big)
\end{aligned}
\label{eq:guard_compose_orig}
\end{equation}
% \vskip -0.1in
where $s$ controls attraction toward the ``positive target'' and 
$r$ controls repulsion from the ``negative target'', i.e. the (noise prediction for the) memorized prompt. 
Notably, GUARD’s ``positive target'' is not an image target; it is a (prompt-)conditional noise prediction that aims to steer the denoising process away from a memorized image and towards an alternative high-quality image.

%Intuitively, the attractive term encourages generation to remain close to a semantically aligned but non-memorized region of the conditional space, while the repulsive term actively pushes the model away from the memorized conditioning direction.

%GUARD is a general framework that admits several instantiations, given by different choices for the positive target $\epsilon_\theta^{+}(x_t, e_p)$. Two main categories of GUARDs are given by: (i) changing the prompt $p$ to a semantically similar prompt $p'$ (e.g. by applying random token replacement or using an external model to obtain a paraphrase) and then setting the positive target as  $\epsilon_\theta(x_t, e_p')$, so that the denoising process fails to generate the original training image that was associated with the  memorized prompt $p$; (ii) changing  $\epsilon_\theta$ itself, e.g. using the method of \citep{ren2024unveiling} that modifies the distribution of cross-attention, aiming to pay less attention to ``trigger tokens'' that are otherwise responsible for generating training data.

%While we find that several of the above GUARDs can effectively mitigate memorization, producing an image that not only is distinct from training data but is also high quality and highly-semantically-related to the given prompt, is a challenging feat that is only possible through a careful design of the positive target. 
%We address this in the next section, where we present a novel choice for $\epsilon^{+}_\theta(x_t, e_p)$ that produces a GUARD that Pareto-dominates all prior methods across all experimental settings.

GUARD allows multiple instantiations through the choice of the positive target $\epsilon_\theta^+(x_t,e_p)$. We next present a concrete novel instantiation. %based on a novel cross-attention spike detection and attenuation approach.

\section{Instantiating the GUARD} % via Cross-Attention Spike Detection and Attenuation}
\label{ref:guard_instantiation_ca}

Section~\ref{sec:ar} introduced GUARD as an abstract inference-time framework.
However, GUARD can only be successful at memorization mitigation without quality degradation if an appropriate positive target is defined that GUARD will steer the generation towards. We therefore devote this section to the pursuit of such a positive target, i.e. an alternative denoising procedure that can produce a high-quality image that is distinct from training data. 
Towards that goal, we revisit a fundamental aspect of generating an image conditional on a text prompt, namely the cross-attention mechanism. We seek to understand the behaviour of the cross-attention distribution for different types of memorization, and accordingly develop a mechanism that adjusts that distribution to prevent memorization while maintaining quality.
Based on this, this section contributes a GUARD by selecting a concrete positive target: a prompt-aligned conditioning signal obtained by attenuating cross-attention (CA) at prompt-specific memorization-critical locations. 
% We now plug the CA spike detector and CA attenuation mechanism into GUARD.

% We now instantiate GUARD by selecting a concrete positive target: a prompt-aligned conditioning signal obtained by attenuating cross-attention (CA) at prompt-specific memorization-critical locations.
% We now plug the CA spike detector and CA attenuation mechanism into GUARD.

% \vskip -0.2in
\subsection{You Don't Need All That (Cross-)Attention}\label{sec:ca_attenuation}

% The fundamental question is: what is it exactly that does not need all that attention? We address this issue here.
Prior work has shown that the main driver of reproducing a training image is the memorization of its associated text prompt; a phenomenon that is tied with disproportionately large attention concentration on specific ``trigger tokens'' that the generation process can latch on to in order to retrieve the original image \citep{wen2024detecting,ren2024unveiling}. 
To address this issue, we therefore seek to modify the attention distribution to prevent memorization triggers from receiving ``all that attention''. But which exactly are the token positions whose attention we should reduce? En route to discovering a fit-for-purpose attention redistribution mechanism, we first contribute an analysis of  the attention patterns of (i) data points that are memorized versus data points that are not memorized, and (ii) for different types of memorization.

% To address this issue, we therefore seek a principled method that redistributes the (cross-)attention in a way that (i) disrupts the model's tendency to place substantial weight on ``memorization trigger tokens'', (ii) is ``surgical'' in that it maintains all other patterns that are needed for the generation to have high semantic alignment with the prompt, and (iii) is ``dynamic'' and adapts the reweighing strategy to the given prompt, since different prompts are associated with different attention patterns and necessitate different treatment. 

% En route to building a cross-attention redistribution mechanism that meets the above desiderata, we start by examining the attention patterns of different data points that are memorized, compared to data points that are not memorized, for different types of memorization.

% The results of this analysis provided the key motivations and justifications for the design of the particular cross-attention redistribution strategy we propose. 
First, as seen in Figure \ref{fig:main-fig-distr-verbatim} for verbatim memorization, the EOT token exhibits a large spike (a finding corroborated in \cite{ren2024unveiling} and leveraged in their mitigation method). In fact, Figure \ref{fig:main-fig-ca-verbatim} shows that, throughout inference steps, the CA mass on the EOT token is much higher for a memorized prompt than for a non-memorized one. However, note that Figure \ref{fig:main-fig-distr-verbatim} shows that several other tokens also exhibit sharp spikes, often surpassing the magnitude of the EOT spike in fact. Thus, going beyond prior findings, we highlight that attenuating only at EOT, without special attention to these other spikes, is not a precise enough intervention.

Second, as seen in Figures \ref{fig:main-fig-distr-template} and \ref{fig:main-fig-ca-template} for template memorization, multiple tokens exhibit pronounced spikes, but unlike the  verbatim case, the cross-attention assigned to EOT in typical memorized examples is no longer higher than that of non-memorized examples (comparing Figure \ref{fig:main-fig-ca-template} with Figure \ref{fig:main-fig-ca-verbatim}). In fact, during the early inference steps, EOT cross-attention is \textit{lower} for memorized examples, with the two converging at later steps. 
This reveals that attenuating CA at EOT, for example, is counter-productive in the case of template memorization, even if it worked well for verbatim memorization. This is corroborated by and explains our evaluation findings whereby the method in \citep{ren2024unveiling} fails to improve or even underperforms prior state-of-the-art methods for template memorization (Table \ref{tab:main_results_shorter}, SD v2.0).

Put together, these findings suggest that there is no fixed rule that describes how cross-attention should be redistributed for memorization mitigation across examples and types of memorization. Therefore, a key desideratum for our attention redistribution strategy is to be dynamic and able to detect locations that require attenuation 
\textit{per prompt}. 

Overall, we seek a principled method that redistributes the (cross-)attention in a way that (i) disrupts the model's tendency to place substantial weight on ``memorization trigger tokens'', (ii) is ``surgical'' in that it maintains all other patterns that are needed for the generation to have high semantic alignment with the prompt, and (iii) is ``dynamic'' and adapts the reweighing strategy to the given prompt, since different prompts are associated with different attention patterns and necessitate different treatment. In the next subsection, we present a method that meets these desiderata.

% , the cross-attention distribution for template memorization evaluated under SD v1.4 for a representative example. 
% Similar to the verbatim memorization case (Figure \ref{fig:ca-distro-nomit}), Figure \ref{fig:ca-distro-nomit-temp} reveals that template memorization is also characterized by multiple memorization trigger tokens that exhibit pronounced CA spikes.

% However, a key difference emerges at the EOT token. As shown by comparing Figure \ref{fig:ca-eot-temp} with Figure \ref{fig:ca-eot}, the cross-attention assigned to EOT in typical memorized examples is no longer higher than that of non-memorized examples. In fact, during the early inference steps, EOT cross-attention is \textit{lower} for memorized examples, with the two converging at later steps. 
% This reveals that attenuating CA at EOT, for example, is counter-productive. This is corroborated by and explains our evaluation findings whereby the method in \citep{ren2024unveiling} fails to improve or even underperforms prior state-of-the-art methods for template memorization (Table 1, SD v2.0).

% From Figure \ref{fig:ca-distro-nomit}, we observe that,. What's more, different training examples spike in different positions \eleni{TODO: add a figure showing this too, as it's important for justifying the adaptivity of our method}.

\subsection{Detecting and Attenuating Attention at Spikes}
\label{sec:detect-and-attenuate}
% \subsection{Locating Spikes}

Our proposed cross-attention redistribution method has two parts: (i) an automatic detection of prompt-specific CA ``spikes'', and (ii) attenuation at those discovered spikes.

\textbf{Locating spikes.} 
We develop an attention-spike detector based on finding statistical outliers in the CA distribution, allowing to locate prompt-specific spikes on-the-fly.  
% Based on these insights, we develop a novel attention attenuation approach that detects statistical outliers in the CA distribution, allowing to find prompt-specific spikes on-the-fly, and subsequently adjusts the distribution of attention to attenuate those locations. 
Concretely, 
% we run the diffusion model conditioned on the target prompt and extract CA maps. This is not straightforward as, for example, not all U-Net blocks can help mitigate memorization \textit{without damaging image quality}. Section \ref{app:block} in the Appendix discusses this in more detail. 
% \todo{add block identification in appendix}).  %from selected U-Net blocks . 
we extract CA distribution maps conditioned under the embedding of the  memorized prompt, $e_p$. We then score token-level spikes as follows: for each token position $i$, we compute the maximum attention mass assigned to that token across spatial queries (aggregating over specific blocks/heads as specified later):
% \vskip -0.1in
$$
M_i \;=\; \max_q \mathrm{AttnScore}_{q,i}.
$$
% \vskip -0.1in
We compute the mean $\mu$ and standard deviation $\sigma$ over $\{M_i\}_i$ and define $Z_i = (M_i-\mu)/\sigma$.
Tokens with $Z_i>\tau$ (e.g.\ $\tau=3$) are flagged as ``spiky''. Let
% \vskip -0.1in
$$
\mathcal{S}(p) \;=\; \{\, i : Z_i>\tau \,\}\ %\cup\ \{ i_{\mathrm{EOT}} \}
$$
% \vskip -0.1in
denote the set of spike positions for prompt $p$ (note that this set can include the EOT position).

\textbf{CA Attenuation at Spikes.}
We next attenuate the detected CA spikes. Concretely, 
% we apply a scaling factor at the logit of each discovered spike location, 
we scale down the corresponding CA logits by a multiplicative factor, thereby suppressing its influence on generation. 
This surgical attenuation yields a modified conditioning signal that remains aligned with the original prompt but with reduced memorization. 
% Delving more into this, we observe in Figures \ref{fig:ca-distro-eot} and \ref{fig:ca-distro-eot-spikes}, that CA attenuation at EOT alone already reduces memorization. And that extending attenuation to additional spike tokens consistently leads to further improvements. 
% In the big picture of the GUARD framework, the resulting conditioning effectively serves as a new positive neighbor. 
%\eleni{just noting: i shortened this paragraph significantly by commenting out outdated things.}

Given a memorized prompt $p$, in each cross-attention layer, we modify the attention logits prior to softmax by 
scaling down the logits corresponding to tokens in $\mathcal{S}(p)$:
% adding a negative bias for tokens in $\mathcal{S}(p)$:
% \vskip -0.2in
$$
\ell_{q,i} \;=\; \frac{\langle Q_q, K_i \rangle}{\sqrt{d}}
\quad \longrightarrow \quad
\ell'_{q,i} \;=\;
% \ell_{q,i} - b \cdot \mathbf{1}[i \in \mathcal{S}(p)]
\ell_{q,i}\cdot \alpha^{\,\mathbf{1}[i \in \mathcal{S}(p)]}
,
$$
% \vskip -0.1in
where $\alpha>0$ controls attenuation strength. 
% \kairan{i recently changed this implementation detail (scaling down CA logits rather than adding negative bias) so correct it here}

We denote the modified noise prediction network obtained by the above attention redistribution mechanism, as $\epsilon^+_\theta(x_t,t,e_p;\mathcal{S}(p),\alpha)$, for a given prompt $p$, which now also depends on the set of automatically-discovered attention spikes $\mathcal{S}(p)$ and the scaling factor $\alpha$, treated as a hyperparameter. For a given prompt $p$, we instantiate GUARD by using $\epsilon^+_\theta(x_t,t,e_p;\mathcal{S}(p),\alpha)$ as the positive target. We refer to this instantiation of GUARD as CA-in-GUARD. We illustrate CA-in-GUARD in a diagram in Figure \ref{fig:guard_diagram}.

Note that by setting $r=0$, CA-in-GUARD reduces to a novel method based solely on CA attenuation, which will be shown to outperform the prior state-of-the-art which leverages
CA attenuation.
%Our evaluation shows that this GUARD ablation outperforms the prior state-of-the-art method based on CA attenuation. % \citep{ren2024unveiling}.}
%\eleni{In our experiments, we will also evaluation our CA attenuation method as a stand-alone method, outside of GUARD, which can be seen as an ablation of GUARD obtained by setting $r=0$, turning off the repulsion term and relying on CFG with our modified forward pass, analogous to \citep{ren2024unveiling}.}

A diagram of the overall inference procedure of CA-in-GUARD is shown in Figure \ref{fig:guard_diagram}

\begin{figure}[t]
\centering
\resizebox{\columnwidth}{!}{%
\begin{tikzpicture}[
    % TIGHT SPACING for single column
    node distance=1.5cm and 0.2cm, 
    box/.style={
        draw, 
        rectangle, 
        rounded corners, 
        align=center, 
        minimum height=1cm, 
        minimum width=2.2cm, 
        fill=white, 
        text width=2.2cm, 
        font=\scriptsize
    },
    greenbox/.style={box, fill=green!10, draw=green!60, thick},
    redbox/.style={box, fill=red!10, draw=red!60, thick},
    spikebox/.style={box, fill=white!10, draw=black!60, thick, text width=0.8cm, minimum width=1.6cm, font=\tiny},
    orangebox/.style={box, fill=orange!10, draw=orange!60, thick, minimum width=7.5cm, text width=7cm},
    guardbox/.style={box, fill=white!10, draw=black!60, thick, text width=4cm, font=\tiny},
    arrow/.style={-Stealth, thick, rounded corners},
    label/.style={font=\scriptsize, align=center, fill=white, inner sep=1.5pt} % Increased padding slightly for better masking
]

    % --- 1. Center Node (Attenuated) ---
    \node[greenbox] (u_pos) at (0,0) {\textbf{Positive target}\\($\epsilon^+$)\\Attenuate at  $\mathcal{S}(p)$};

    % --- 2. Left and Right Nodes ---
    \node[redbox, left=0.2cm of u_pos] (u_neg) {\textbf{Negative target}\\($\epsilon^-$)\\Standard CA on $\epsilon_p$};
    \node[box, right=0.2cm of u_pos] (u_unc) {\textbf{Unconditional}\\($\epsilon_\phi$)\\Standard CA on $\epsilon_\phi$};

    % --- 3. Input Node ---
    \node[box, above=2cm of u_pos] (inputs) {\textbf{Inputs}\\Prompt ($p$), partially-desnoised image $x_t$};

    % --- 4. Spike Detector ---
    \path (u_neg.north) -- (u_pos.north) coordinate[midway] (mid_branch);
    \node[spikebox, above=0.42cm of mid_branch] (detect) {\textbf{Spike Detector}\\Output $\mathcal{S}(p)$};

    % --- 5. Combination Node ---
    \node[guardbox, below=0.8cm of u_pos, minimum height=0.8cm] (combination) {$\hat{\epsilon} = \epsilon_{\phi} + s(\epsilon^{+} - \epsilon_{\phi}) - r(\epsilon^{-} - \epsilon_{\phi})$};

    % --- 6. Final Output ---
    \node[box, below=0.4cm of combination, minimum width=3cm, text width=3cm, minimum height=0.6cm] (final) {\textbf{Final Prediction} ($x_{t-1}$)};

    % --- ARROWS ---
    
    % 1. Left Branch (Memorized)
    \draw[arrow] (inputs.south) -- ++(0,-0.3) -| (u_neg.north);

    % 2. Right Branch (Unconditional)
    \draw[arrow] (inputs.south) -- ++(0,-0.3) -| (u_unc.north);

    % 3. Center Branch (Main Feed)
    \draw[arrow] (inputs.south) -- (u_pos.north);

    % 4. Spike Detector Logic
    % FIX: Start from 'north west' to avoid overlapping the center label
    % \draw[arrow, dashed] (u_neg.north west) |- (detect.west);
    
    % Injection (Solid)
    \coordinate (merge_point) at (u_pos.north |- detect.east);
    \draw[arrow] (detect.east) -- (merge_point);

    % 5. Outputs to Combination
    \draw[arrow] (u_neg.south) -- ++(0,-0.3) -| (combination.north);
    \draw[arrow] (u_pos.south) -- (combination.north);
    \draw[arrow] (u_unc.south) -- ++(0,-0.3) -| (combination.north);

    % 6. Final
    \draw[arrow] (combination) -- (final);

    % --- LABELS (Drawn LAST to ensure they sit on top) ---
    % Height definition
    \coordinate (label_y) at ($(u_pos.north) + (0, 0.4)$);
    
    % Place the labels
    \node[label] at (u_neg.north |- label_y) {$\epsilon_p$};       % Left Box
    \node[label] at (u_pos.north |- label_y) {$\epsilon_p$};       % Middle Box
    \node[label] at (u_unc.north |- label_y) {$\epsilon_\emptyset$}; % Right Box

\end{tikzpicture}%
}
\caption{Overview of the CA-in-GUARD denoising process.}
\label{fig:guard_diagram}
\end{figure}

\subsection{Overall Inference-Time Procedure}
\label{sec:procedure}
For a given prompt $p$ and at each denoising step $t$, the simplest way to think of the proposed method is that it performs three U-Net evaluations:
(1) an \textbf{unconditional} pass to obtain $\epsilon_\theta(x_t,t,e_\phi)$;
(2) a \textbf{memorized conditional} pass (with standard CA) to obtain $\epsilon^-_\theta(x_t,t,e_p)$ and to extract CA maps for spike detection ;
(3) a \textbf{spike-attenuated conditional} pass (same $e_p$, but with CA logits attenuated at $\mathcal{S}(p)$) to obtain $\epsilon^+_\theta(x_t,t,e_p;\mathcal{S}(p),\alpha)$.
The final prediction $\hat{\epsilon}(x_t,t)$ is then formed by Eq.\,(\ref{eq:guard_compose_orig}) and used to obtain $x_{t-1}$.
In practice, we compute the three noise predictions in a single forward pass of the U-Net per timestep, by batching the null prompt, the positive neighbor conditioning, and the memorized prompt together, yielding efficiency close to that of a single forward pass; see Section \ref{sec:implementation_details} for more details. 

\textbf{How $\mathcal{S}(p)$ is computed.}
We compute spike scores $Z_i$ from the per-prompt CA distributions extracted under the memorized conditional pass, using $M_i=\max_q \mathrm{AttnScore}_{q,i}$ and thresholding $Z_i>\tau$, then 
set $\mathcal{S}(p)=\{i:Z_i>\tau\} $.

\textbf{When $\mathcal{S}(p)$ is computed.}
We compute $\mathcal{S}(p)$ at every diffusion step and apply attenuation adaptively until no spikes are detected. This continuous strategy allows us to capture and suppress both early and late-stage attention spikes. 
% Although $\mathcal{S}(p)$ could be fixed after early-to-mid timesteps to reduce overhead, we found continuous computation to be more stable in practice.

% To reduce overhead, $\mathcal{S}(p)$ can be computed on a small set of \emph{early-to-mid} steps (e.g.\ the first $T_{\mathrm{detect}}$ steps) and then \emph{held fixed} for the remaining steps. 
% Alternatively, it can be recomputed every $K$ steps; in our experiments, fixing $\mathcal{S}(p)$ after early-to-mid detection achieves comparable performance with lower variance.

\textbf{Which blocks/heads are modified.}
CA-logit attenuation is applied only in selected cross-attention modules of the U-Net. By default, we apply it in the \emph{down} and \emph{mid} blocks, 
, while omitting late \emph{up} blocks to avoid quality degradation.
% , such as texture artifacts.
% We omit late \emph{up}-block attenuation to avoid reducing quality such as introducing texture artifacts. 
Within each selected module, attenuation can be applied either to all heads, or restricted to \emph{hot heads} (heads that allocate unusually large mass to $\mathcal{S}(p)$). 
We found that applying attenuation to all heads is a strong and simple default.
Section \ref{app:block} in the Appendix discusses this in more detail.

\section{Experimental Protocol}
\label{sec:protocol}
This section describes our evaluation protocol, outlining and justifying important differences over prior work. 
We depart from the protocol used in prior work in two important ways. First, we exclude examples that have relatively low memorization scores to begin with, as those examples are easy for all methods, so their inclusion may hide inherent problem difficulties and overestimate achievable performance.
%dilutes the overall results. 
Second, we report results separately for verbatim vs template memorization, which illuminates previously-undiscovered issues in performance of certain methods. 

% \subsection{Experimental setup}
\textbf{SD models and datasets.} 
We conduct experiments on the dataset of 500 memorized samples identified by \citet{webster2023reproducible}, which contains both verbatim and template memorization cases. 
% We report results separately for the verbatim and template subsets. %under their corresponding memorization settings.
Our experiments use Stable Diffusion (SD) v1.4 (pretrained on LAION-2B \citep{schuhmann2022laion}) and SD v2.0 (pretrained on a deduplicated subset of LAION-5B \citep{schuhmann2022laion}). 
Because verbatim memorization is much rarer in newer models pretrained on deduplicated data \citep{webster2023reproducible}, we study both verbatim and template memorization on the earlier SD v1.4, but focus on template memorization only for SD v2.0. 

\textbf{Metrics.} We quantify memorization using the SSCD  score that measures the similarity between the image generated with a given prompt compared to the original training image that was associated with that prompt \citep{pizzi2022self}, where higher SSCD indicates stronger memorization.  
We measure generation quality using CLIP score \citep{radford2021learning} (higher indicates better prompt-image alignment) and FID \citep{FID_heusel2017gans} (lower indicates better realism and diversity). We compute FID between images generated from the memorized prompts and a reference set of 10,000 images sampled from LAION \citep{schuhmann2022laion}.
% Additionally, we include the mean CLIP score of LAION-168k as a reference for ``good quality'' generations that we ideally would obtain for the memorized examples.

\textbf{Memorized prompt selection.} Unlike recent work \citep{han2025adjusting}, we further filter memorized prompts and retain only highly memorized examples,
% : We only consider actual highly memorized ones, 
i.e., having an SSCD score $>0.7$. Our results reveal that omitting this filtering can hide the difficulties associated with memorization mitigation and provide a false sense of safety. Averaging SSCD scores over both highly and mildly memorized examples can artificially inflate the memorization mitigation performance of algorithms, which otherwise would struggle. This is clearly evident in template memorization and SD v2.0, where many previously identified memorized examples exhibit low SSCD scores. Hence, it is unsurprising that \citet{han2025adjusting} (and the other baselines) reports substantially better SSCD performance (lower SSCD values), as their evaluation starts from a lower memorization baseline.
% is shown to be considerably higher (i.e. they can achieve lower SSCD scores, since they start from a lower bar).

\textbf{Baselines.}
We evaluate our method against prior state-of-the-art baselines, including Random Token Addition (RTA) \citep{somepalli2023understanding},  \citet{wen2024detecting}, \citet{ren2024unveiling}, \citet{han2025adjusting}, and the original model without any memorization mitigation as the reference point,
% , using their recommended settings whenever applicable
to assess their ability to mitigate the memorization via an inference-time intervention.

\textbf{Hyperparameter selection.} 
% \label{sec:hyperparam-selection}
There is a multitude of diverse criteria and trade-offs that we must take into consideration.
%In evaluating the success of a memorization mitigation method, there are diverse criteria and trade-offs that we must take into consideration.
% Aside from achieving lower SSCD (better memorization), we also care about the quality of generations, both in terms of semantic adherence to the prompt and perceptual quality. 
% We are thus in need of a principled protocol to establish an appropriate, comprehensive and fair comparison setup that takes all relevant trade-offs into account. 
We therefore adopt a principled selection protocol to ensure a fair and comprehensive comparison.
% a principled manner in reporting performance taking into account all relevant trade-offs.
% Computational constraints make exhaustive exploration of all hyperparameter combinations of all baselines and for all metrics (SSCD, CLIP, FID) prohibitively expensive. 
% Instead, we opted to adopt a principled   protocol to establish a comprehensive and fair comparison setup.
% To this end, 
We first run SD v1.4 and v2.0 on a subset of their pretraining data to obtain reference CLIP scores 
that represent strong image generation quality, which we aim to preserve.
% representing a strong reference point for the image generation quality, which we want to preserve. 
The rationale is that 
although some methods may substantially reduce memorization, this is of no practical use if image quality degrades severely.
% generated image quality is very bad.
Said reference points were found to be 0.299 for SD v1.4 and 0.323 for SD v2.0. 
We allow up to a 15\% degradation from these references, providing flexibility to identify configurations that trade moderate quality loss for strong memorization mitigation.
% We then deemed as acceptable and allowed up to a 15\% CLIP degradation from the reference point.
% This induces flexibility to detect cases where although quality is worsened, methods may achieve very strong memorization mitigation. 
Among runs within this acceptable range, we report the best results for each method in terms of SSCD, CLIP, and FID.
% Finally, among experiment runs within this range, we report the best results in terms of SSCD, CLIP and FID, respectively, for each method. 
For SD v1.4 with verbatim memorization, this yields a minimum acceptable CLIP score of 0.254. For template memorization, where all methods (including no mitigation) are far below the pretraining reference, we instead use the no-mitigation CLIP score as the reference and again allow a 15\% degradation, yielding minimum acceptable CLIP scores of 0.186 for SD v1.4 and 0.183 for SD v2.0. 
% In the next section, when we say we select the configuration that achieves the best CLIP, we refer to the best CLIP within this acceptable range.
Throughout the next section, ``best CLIP'' refers to the highest CLIP score achieved within these acceptable ranges.

For completeness, we also report results obtained using the traditional evaluation protocol of prior baselines,
% Nonetheless, we have also run all experiments using the protocol of all prior baselines 
which does not apply this hyperparameter selection strategy. These results are shown in Figure \ref{fig:pareto-all} in Section \ref{app:all-methods}.

\begin{table*}[t]
\centering
\caption{Evaluation results across metrics, architectures and memorization types. For each setting (architecture and memorization type), we select for each method the configuration that yields the best SSCD, since this is the primary metric for memorization mitigation (in Figure \ref{fig:standalone_example}, and in more detail in Table \ref{tab:main_results} in the Appendix, we show results with different selection criteria too). For each prompt, we generate four images and report the mean $\ci$ (95\% confidence interval) across four generations.
% \kairan{
Additional evaluation results on SD v3.0 are reported in Table \ref{tab:sdv3-results} in \ref{app:sdv3-evaluation}.
% }
}

% =========================
% Best SSCD
% =========================
\footnotesize
\resizebox{\textwidth}{!}{%
\begin{tabular}{l|ccc|ccc|ccc}
\toprule
 & \multicolumn{3}{c|}{SD v1.4 -- verbatim memorization}
 & \multicolumn{3}{c|}{SD v1.4 -- template memorization}
 & \multicolumn{3}{c}{SD v2.0 -- template memorization} \\
Method
 & SSCD ($\downarrow$) & CLIP ($\uparrow$) & FID ($\downarrow$)
 & SSCD ($\downarrow$) & CLIP ($\uparrow$) & FID ($\downarrow$)
 & SSCD ($\downarrow$) & CLIP ($\uparrow$) & FID ($\downarrow$)\\
\midrule

No mitigation
 & \ci{0.875}{0.001} & \ci{0.346}{0.001} & 243.056
 & \ci{0.776}{0.017} & \ci{0.219}{0.007} & 258.976
 & \ci{0.735}{0.011} & \ci{0.215}{0.005} & 303.266 \\

RTA
 & \ci{0.328}{0.007} & \ci{0.263}{0.002} & 175.866
 & \ci{0.617}{0.043} & \ci{0.187}{0.010} & 218.343
 & \ci{0.543}{0.048} & \ci{0.183}{0.009} & 233.580 \\

Wen et al.
 & \ci{0.115}{0.011} & \ci{0.267}{0.003} & 162.848
 & \ci{0.545}{0.038} & \ci{0.188}{0.008} & 209.719
 & \ci{0.260}{0.026} & \ci{0.183}{0.008} & 188.914 \\

Ren et al.
 & \ci{0.113}{0.007} & \ci{0.258}{0.005} & 164.638
 & \ci{0.602}{0.033} & \ci{0.184}{0.007} & 222.066
 & \ci{0.356}{0.024} & \ci{0.188}{0.007} & 208.416 \\

Han et al.
 & \ci{0.191}{0.016} & \ci{0.256}{0.008} & 166.551
 & \ci{0.479}{0.033} & \ci{0.188}{0.006} & 210.839
 & \ci{0.401}{0.024} & \ci{0.186}{0.005} & 208.852 \\

\midrule

CA attenuation
 & \ci{0.109}{0.006} & \ci{0.282}{0.004} & 164.660
 & \ci{0.530}{0.038} & \ci{0.185}{0.009} & 212.240
 & \ci{0.193}{0.014} & \ci{0.184}{0.005} & 245.850 \\

CA-in-GUARD
 & \ci{0.079}{0.007} & \ci{0.266}{0.015} & 158.115
 & \ci{0.517}{0.038} & \ci{0.186}{0.008} & 210.983
 % & \ci{0.223}{0.017} & \ci{0.183}{0.006} & 187.250
  % & \ci{0.193}{0.014} & \ci{0.183}{0.005} & 212.727
 & \ci{0.193}{0.014} & \ci{0.183}{0.005} & 212.727
 \\

\bottomrule
\end{tabular}
}

\label{tab:main_results_shorter}
\end{table*}

\begin{figure*}[t]
\centering
\begin{subfigure}[b]{0.34\textwidth}
 \centering
\includegraphics[scale=0.3]{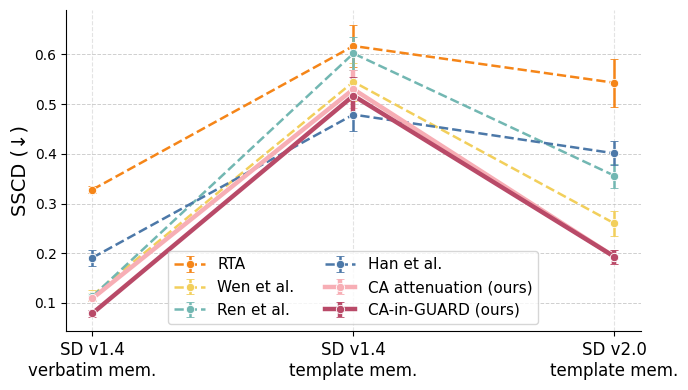}
 \caption{Best SSCD ($\downarrow$)}
 \label{fig:itmm-sscd}
\end{subfigure}
%  \hfill
\begin{subfigure}[b]{0.34\textwidth}
\centering
\includegraphics[scale=0.3]{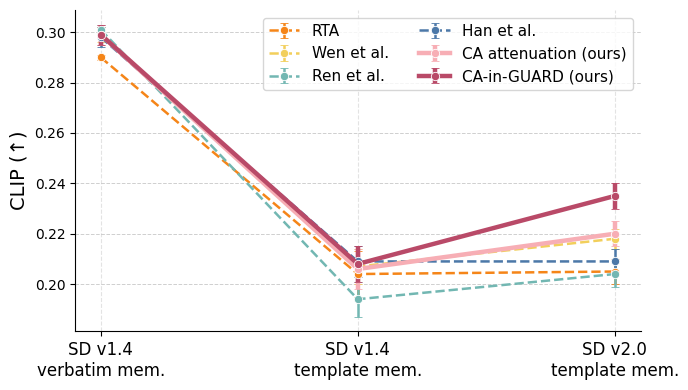}
\caption{Best CLIP ($\uparrow$)}
\label{fig:itmm-clip}
\end{subfigure}
%  \hfill
\begin{subfigure}[b]{0.3\textwidth}
\centering
\includegraphics[scale=0.3]{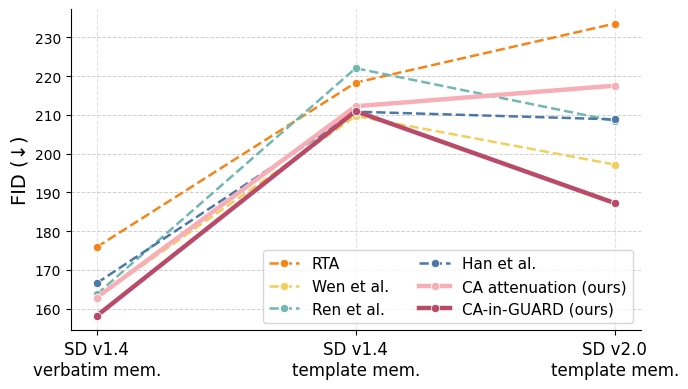}
\caption{Best FID ($\downarrow$)}
\label{fig:itmm-fid2}
\end{subfigure}
\caption{
\textbf{The \textit{best achievable} SSCD, CLIP, and FID of different methods}. We plot the best value a method can achieve on each metric \textit{individually}, using the configuration that yields best results on that specific metric.
In each subplot, a different configuration may be used (we pick the hyperparameter setting that yields the best SSCD, best CLIP and best FID, respectively), so this plot does not speak to the ability of a method to do well on all metrics \textit{jointly}, nor to trade-offs between these metrics, which we investigate separately.
% Performance across 3 settings: SD v1.4 with verbatim memorization, template memorization, and SD v2.0 with template memorization (x-axis). 
% Each subplot reports a single metric (y-axis) and represents the best metric value these methods can achieve: SSCD (lower is better), CLIP (higher is better), or FID (lower is better).
% ``Best of baselines'' denotes the best-performing baseline per setting (selected post hoc among the baselines) for comparison against GUARD.
}
\label{fig:best_achievable}
\end{figure*}

\section{Experimental Results}
\label{sec:results}
%In our view, a good 
Memorization mitigation must satisfy two criteria. First, %it is able to 
achieve strong results in terms of any given metric, allowing a developer to configure it to achieve best results for their prioritized metric, for their specific use case. Second, it %costs 
must incur minimal degradation in other metrics of interest when configured to work as well as possible for a given metric.  
%Based on this, 
We thus analyze the performance of methods in two ways: (i) we investigate their best achievable SSCD, CLIP, and FID, where the ``best achievable'' value for a metric is obtained by selecting the best configuration for that metric individually (in Figure \ref{fig:best_achievable}), and then (ii) we explore what is the effect on SSCD, a primary metric, when configuring for achieving the best CLIP (in Figure \ref{fig:standalone_example}, and in full detail in the Appendix Table \ref{tab:main_results}) and what is the effect on CLIP and FID metrics metrics when selecting the best configuration for SSCD (in Table \ref{tab:main_results_shorter} and in full detail in Table \ref{tab:main_results}), thus illuminating methods' trade-offs between different metrics. We describe our main findings below. 

% \subsection{Does our CA attenuation method improve over prior CA intervention methods?}
\textbf{Our CA attenuation alone outperforms prior work}.
We begin by comparing our CA attenuation method against the previous state-of-the-art method of  
\citet{ren2024unveiling}, which is also based on CA attenuation and is thus is the most closely related prior method. %\eleni{hmm, should we say that this is SOTA too? (if it is?) or that we also outperform prior work too?} 
%\peter{I think this is enough... Ren et al is not sota, even Wen et al outperforms it ... we could say one of the prior SOTA?}
% Both approaches intervene at inference time by manipulating cross-attention to mitigate memorization. \citet{ren2024unveiling} apply a fixed and token-specific intervention by setting the CA of the EOT token to $-\infty$ and inflating the CA of the beginning-of-text (BOT) token. 
% In contrast, our method does not assume memorization is tied to a single special token, and instead identifies and attenuates CA spikes more generally, directly addressing the need for a finer-grained and  per-prompt adaptive spike detection method that we identified in Section \ref{sec:ca_attenuation}.
% Figure \ref{fig:ar-ca-eccv} and 
Figure \ref{fig:best_achievable} and Table \ref{tab:main_results_shorter} show that our CA attenuation method alone significantly improves upon \citet{ren2024unveiling} in terms of mitigating memorization, as shown by the SSCD metric, \textit{across all settings} of architecture version and memorization type.
The gap is particularly large especially for template memorization, and especially for SD v2.0. Concretely, our CA attenuation yields an SSCD of 0.53 (vs 0.60 for \citet{ren2024unveiling}) for template memorization on SD v1.4, and 0.19 (vs 0.36 for \citet{ren2024unveiling} for template memorization on SD v2.0, which are drastic gaps. This large improvement stems directly from our analysis on CA distribution and the findings (Figure \ref{fig:all-motivating-figs}) that attenuating CA only at EOT is insufficient.

% Figure \ref{fig:sscd-across-configs} also corroborates the ability of our CA attenuation to yield significantly better SSCD than all prior work. Notably, when selecting the configuration that is best for CLIP, our CA attenuation method yields a much lower SSCD relative to the SSCD obtained by \cite{ren2024unveiling}'s method when configuring it for best CLIP (). 

The improved memorization mitigation of our CA attenuation over prior methods for CA attenuation, however, comes at some cost of quality on the FID metric (see Figure \ref{fig:itmm-fid2}), while CLIP is comparable. Figure \ref{fig:ar-ca-eccv} in Section \ref{app:ar-ca-eccv} shows that we can fully recover and boost quality significantly beyond prior work by incorporating this method into GUARD.

% \begin{figure}[h]
% \centering
% % \begin{subfigure}[b]{0.34\textwidth}
% %  \centering
% \includegraphics[scale=0.35]{plots/ca_eccv_sscd.png}
%  % \caption{Best SSCD}
%  % \label{fig:itmm-sscd}
% % \end{subfigure}
% \caption{Performance across 3 settings: SD v1.4 with verbatim memorization, template memorization, and SD v2.0 with template memorization. 
% We report the best \textit{SSCD} achieved by each method. For results for CLIP and FID, see the Appendix.}
% \label{fig:ca-eccv}
% \end{figure}

% Moreover, we note that GUARD's \eleni{framework vs instance} \peter{when we say GUARD's CA attenuation, its clear tha this is the instantiated version, no?} \eleni{hmm but this section is about plain CA} CA attenuation dramatically improves SSCD for template memorization, especially, SD v2.0. This stems directly from our analysis on CA distribution and the findings (Figure \ref{fig:all-motivating-figs}) that attenuating CA only at EOT is insufficient.
%We attribute this gap to the difference between verbatim and template memorization. While suppressing EOT attention is effective for verbatim memorization in SD v1.4, it becomes much less effective for template memorization. 
%As shown in Figure \ref{fig:ca-distro-temp}, in the template setting the EOT token does not attract strong CA, whereas other tokens still produce pronounced attention spikes. Consequently, EOT-specific attenuation is insufficient, while our spike-based CA attenuation remains effective by directly targeting the dominant attention locations.

\textbf{CA-in-GUARD improves even more.} Having established that our CA attenuation alone improves upon the prior state-of-the-art for SSCD, our next finding is that we can gain further improvements by encasing our CA attenuation as the positive target of the GUARD framework, yielding the CA-in-GUARD method. Specifically, Figure \ref{fig:best_achievable} reveals that, compared to plain CA attenuation, CA-in-GUARD can further improve (or maintain equally good) performance
%or maintain roughly the same best achievable value as plain CA attenuation, 
for every single metric and setting. 
We note in particular an improvement in SSCD (SD v1.4, verbatim memorization) and large improvements 
%especially 
in terms of CLIP and FID (especially in SD v2.0), mitigating the drawback of plain CA attenuation mentioned above. 
We hypothesize that allowing the repulsion term of GUARD to also play a part in memorization mitigation can alleviate the pressure of strict CA spike reduction, which may come at a cost of quality, if acting as the sole memorization mitigator. This observation suggests that the repulsion term of GUARD acts synergistically with the spike attention attenuation mechanism to yield good trade-offs between mitigating memorization while preserving quality 
({Figure \ref{fig:ar-ca-eccv} in Section \ref{app:ar-ca-eccv}). %for more focused comparison results, including only the relevant methods: Ren et al. vs CA attenuation vs CA-in-GUARD}

\begin{figure}[t]
    \centering
    % Use \columnwidth to ensure it fills the single column exactly
    \includegraphics[scale=0.3]{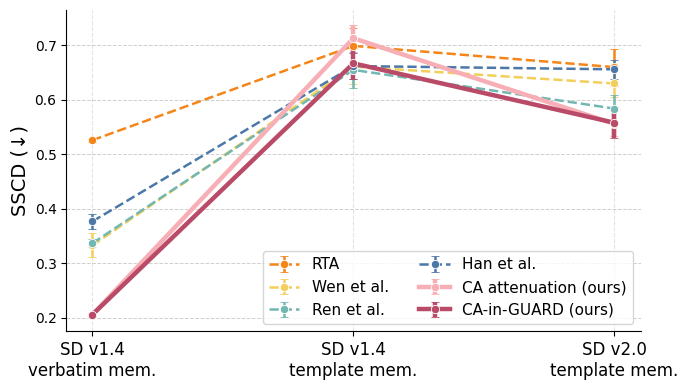}
    \caption{\textbf{Comparison of the SSCD of different methods \textit{for the hyperparameter configuration that works best for CLIP}}. 
    % Our methods achieve by far the best SSCD in this setting too, showing their ability to perform strongly jointly across metrics.}
    }
    \label{fig:standalone_example}
\end{figure}

\begin{figure*}[htb]
\centering
\begin{subfigure}[htb]{\textwidth}
 \centering
\includegraphics[scale=0.3]{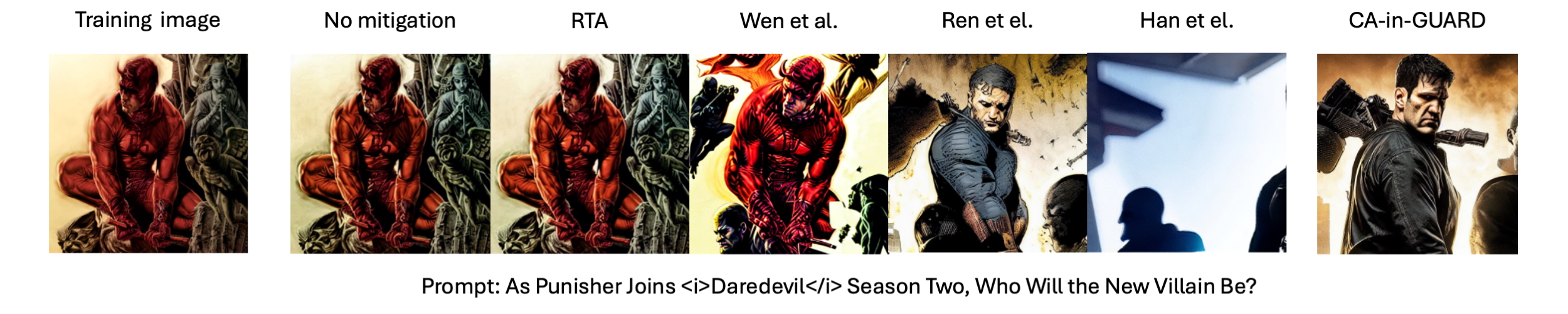}
 % \caption{SD v1.4 - verbatim memorization}
 % \label{fig:itmm-sscd}
\end{subfigure}
%  \hfill
\begin{subfigure}[htb]{\textwidth}
\centering
\includegraphics[scale=0.3]{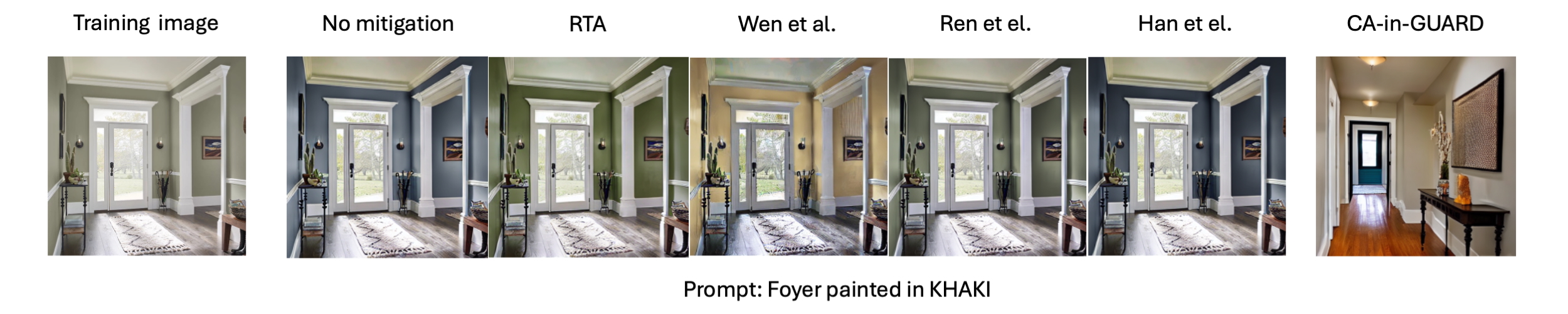}
% \caption{SD v1.4 - template memorization}
% \label{fig:itmm-clip}
\end{subfigure}
%  \hfill
\caption{Qualitative examples on SD v1.4 under verbatim (top row) and template memorization (bottom row). 
No mitigation often reproduces the training example closely. 
CA-in-GUARD can significantly mitigate memorization while preserving prompt-relevant content and image quality. 
Additional examples are provided in \ref{app:qualitative-examples}.}
\label{fig:qualitative-main}
\end{figure*}

\textbf{CA-in-GUARD dominates across settings}. A key overarching conclusion is that the state-of-the-art among all prior research is not a single method, but changes depending on the setting.
For example, Table \ref{tab:main_results_shorter} shows that for SD v1.4 template memorization, \citet{han2025adjusting} is 
the strongest performer for SSCD (0.479) out of prior work (excluding our methods). However, \citet{han2025adjusting} is outperformed by \citet{wen2024detecting} and \citet{ren2024unveiling} for template memorization on SD v2.0, for example. 
%(However, its confidence interval overlaps with that of CA-in-GUARD and that CA-in-GUARD is dramatically better than \citep{han2025adjusting} for SD v1.4 verbatim memorization (0.079 vs 0.191) and for SD v2.0 (0.193 vs 0.401).} 
% Between \citet{wen2024detecting} and \citet{ren2024unveiling}, the former significantly outperforms the latter on template memorization (for both SD v1.4 and SD v2.0) but the latter outperforms the former for verbatim memorization. \peter{NO. Table 1 shows same perf for wen and ren}
Generally, no prior method performs robustly across settings.  
% However, for SD v1.4 verbatim, \citep{wen2024detecting} and \citep{ren2024unveiling} are the strongest among prior work.
On the other hand, CA-in-GUARD provides overall better performance across all settings considered, both when looking at its per-metric best achievable results relative to the per-metric best achievable results of other methods (Figure \ref{fig:best_achievable}), as well as its trade-offs across metrics: Figure \ref{fig:standalone_example} shows that CA-in-GUARD by far improves on the SSCD of prior methods, even when configuring for best CLIP, and Table \ref{tab:main_results_shorter} shows that, when all methods are configured for best SSCD, CA-in-GUARD outperforms prior methods on nearly all settings in terms of SSCD and sometimes in terms of CLIP and FID too, otherwise yielding competitive but not improved results on the quality metrics while improving SSCD. 
% \kairan{
Additional evaluations further support this conclusion: CA-in-GUARD remains robust across samplers, step counts, guidance schedules, and CFG scales (Section \ref{app:ca-in-guard-ablations}); 
remains top-performing under DINO-based retrieval metrics \citep{caron2021emerging}(Sections \ref{sec:sscd-dino-evaluation});
and continues to achieve the strongest overall SSCD results over the full 500-example spectrum, rather than only the high-memorization subset (Section \ref{sec:full-spectrum-evaluation}). 
Qualitative examples in Figure \ref{fig:qualitative-main} (and Section \ref{app:qualitative-examples}) show the same trend visually, with CA-in-GUARD moving generations away from memorized training images while preserving prompt-relevant content.
Moreover, we go beyond the standard prior-work setting on SD v1.4 and SD v2.0 by reporting an evaluation on SD v3.0 in Section \ref{app:sdv3-evaluation}, which, to our knowledge, presents the first evaluation of any memorization mitigation methods on SD v3.0. 
CA-in-GUARD continues to outperform the applicable baselines on this newer architecture, suggesting that its empirical gains extend beyond the SD v1.4/v2.0 setting.
% outperforms applicable baselines in our SD v3.0 evaluation, which to our knowledge is the first evaluation of memorization mitigation methods on SD v3.0 (Section \ref{app:sdv3-evaluation}).
% This conclusion is further supported by the additional evaluations in Section \ref{app:ca-in-guard-ablations}, \ref{sec:sscd-dino-evaluation},  and \ref{sec:full-spectrum-evaluation}: CA-in-GUARD remains robust across common inference choices including samplers, denoising step counts, guidance schedules, and CFG scales;
% remains top-performing under DINO-based retrieval metrics \citep{caron2021emerging};
% and continues to achieve the strongest overall SSCD results when evaluated over the full 500-example spectrum (rather than only the high-memorization subset).
% }
% \kairan{While our main evaluation follows the standard prior-work setting on SD v1.4 and SD v2.0, we additionally report an evaluation on SD v3.0 in Section \ref{app:sdv3-evaluation}, which, to our knowledge, presents the first evaluation of any memorization mitigation methods on SD v3.0. 
% CA-in-GUARD continues to outperform the applicable baselines on this newer architecture, suggesting that its empirical gains are robust beyond the SD v1.4/v2.0 setting.
% }
Generally, CA-in-GUARD is by far the most consistent method in terms of performing strongly across metrics and settings.

% \kairan{
% \textbf{Qualitative examples.}
% Figure \ref{fig:qualitative-main} provides qualitative examples for SD v1.4 under both verbatim and template memorization. Without mitigation, the generated images closely reproduce the corresponding training examples. Prior mitigation methods reduce memorization to varying degrees, but often introduce visible artifacts, semantic drift, or reduced prompt fidelity. In contrast, CA-in-GUARD substantially changes the generated image away from the memorized training example while preserving the main prompt semantics and visual quality. 
% These examples visually corroborate the quantitative trends in Table \ref{tab:main_results_shorter}: CA-in-GUARD provides strong memorization mitigation while maintaining competitive generation quality.
% }

\textbf{Mitigating template memorization is harder.} 
% The main results are shown in Figure \ref{fig:itmm} and Table \ref{tab:itmm}.
Across all baselines, template memorization consistently proves more challenging to mitigate than verbatim memorization. 
While the ``no mitigation'' reference point exhibits lower SSCD under template memorization (which indicates a weaker initial memorization signal), all baseline methods degrade substantially when moving from verbatim to template memorization, with worse SSCD, CLIP, and FID scores. 
This degradation persists in both SD v1.4 and SD v2.0, suggesting that template memorization is a more structurally challenging problem for existing mitigation approaches. 
Figure \ref{fig:best_achievable}, \ref{fig:standalone_example} and Table \ref{tab:main_results_shorter} 
% and \ref{fig:standalone_example} 
show that, across settings, CA-in-GUARD consistently outperforms prior methods on template memorization, yielding a large reduction in SSCD in many cases while having comparable CLIP and FID. 
% \peter{BUT table 4 is the runtimes... you mean table 3?}
% \kairan{i guess it should be figure 4 instead of table 3..? changed it figure 4.}
% Compared to each individual baseline, GUARD is either clearly better or statistically comparable in terms of SSCD, CLIP, and FID. 
% We discuss these results in more detail for each setting below.
% }

\textbf{Other analyses.} We also explore the efficiency of our method (Section \ref{app:runtime}),
other GUARD instantiations (Section \ref{app:guard-alternative-target}), and which U-Net blocks, heads, and denoising steps to operate on
(Sections \ref{app:block}, \ref{app:head}, \ref{app:step}, respectively).

\section{Conclusion} 
We have proposed GUARD, a new framework for inference-time memorization mitigation in T2I diffusion models, and a concrete instantiation of it, CA-in-GUARD, based on an improved CA attenuation method that we developed, motivated by our empirical analyses. Our CA attenuation approach alone improves upon prior state-of-the-art in terms of memorization mitigation, and its integration with GUARD yields by far the most robust method in terms of consistently producing strong memorization mitigation results, across architectures (SD v1.4, v2.0, and v3.0) and memorization types (verbatim and template memorization), while also improving or remaining competitive with prior methods on image quality. We hope future work develops even better instantiations of GUARD, and investigates incorporating the fundamentals of our proposal into training-time or finetuning-time mitigation approaches.

\section*{Acknowledgements}
We thank Jamie Hayes for useful feedback on an earlier draft of this work.

% \textbf{Do not} include acknowledgements in the initial version of the paper

% If a paper is accepted, the final camera-ready version can (and usually should)
% include acknowledgements.  Such acknowledgements should be placed at the end of
% the section, in an unnumbered section that does not count towards the paper
% page limit. Typically, this will include thanks to reviewers who gave useful
% comments, to colleagues who contributed to the ideas, and to funding agencies
% and corporate sponsors that provided financial support.

\section*{Impact Statement}
This work explores the problem of memorization mitigation in text-to-image (T2I) diffusion models, with possible implications for improving privacy, copyright compliance, and responsible model deployment. As T2I systems become increasingly integrated into real-world applications, the ability to suppress the memorization of specific training examples is critical for addressing ethical,  legal, and societal concerns.  
% Authors are \textbf{required} to include a statement of the potential broader
% impact of their work, including its ethical aspects and future societal
% consequences. This statement should be in an unnumbered section at the end of
% the paper (co-located with Acknowledgements -- the two may appear in either
% order, but both must be before References), and does not count toward the paper
% page limit. In many cases, where the ethical impacts and expected societal
% implications are those that are well established when advancing the field of
% Machine Learning, substantial discussion is not required, and a simple
% statement such as the following will suffice:

% ``This paper presents work whose goal is to advance the field of Machine
% Learning. There are many potential societal consequences of our work, none
% which we feel must be specifically highlighted here.''

% The above statement can be used verbatim in such cases, but we encourage
% authors to think about whether there is content which does warrant further
% discussion, as this statement will be apparent if the paper is later flagged
% for ethics review.

% % In the unusual situation where you want a paper to appear in the
% % references without citing it in the main text, use \nocite

% \bibliographystyle{plain}
% \bibliographystyle{plainnat}
\bibliography{references}
\bibliographystyle{icml2026}

%%%%%%%%%%%%%%%%%%%%%%%%%%%%%%%%%%%%%%%%%%%%%%%%%%%%%%%%%%%%
\clearpage

\appendix

\section{Appendix}
% Technical appendices with additional results, figures, graphs and proofs may be submitted with the paper submission before the full submission deadline (see above), or as a separate PDF in the ZIP file below before the supplementary material deadline. There is no page limit for the technical appendices.
% \tableofcontents
\etocsettocdepth{subsubsection}
\localtableofcontents

% \subsection{Impact Statement \& Limitations}
\subsection{Limitations}

% \textbf{Impact Statement.} 
% This work explores the problem of memorization mitigation in text-to-image (T2I) diffusion models, with possible implications for improving privacy, copyright compliance, and responsible model deployment. As T2I systems become increasingly integrated into real-world applications, the ability to suppress the memorization of specific training examples is critical for addressing ethical,  legal, and societal concerns.  
% The proposed framework is general and may inform future developments across other generative tasks, including image-to-image generation, video generation, and multimodal generation, extending the impact beyond text-conditioned models.

% \textbf{Limitations.} 
% \eleni{this was outdated and no longer accurate, i tried rewriting it.}
A limitation of our method, and indeed any inference-time memorization mitigation approach, is that it does not (even attempt to) erase memorized information from the model weights. It simply attempts to prevent it from overly affecting the generation process. While we operate under this framework by choice, to avoid other issues inherent in training-time and finetuning-time mitigation approaches, we acknowledge that the family of inference-time methods has the inherent limitation of not being able to protect against any threat model (e.g. an adversary that has white-box access to the weights can still retrieve the memorized information). Nonetheless, we believe that inference-only mitigation methods can be practically very important, under realistic black-box threat models, and due to their promise to yield better trade-offs in terms of efficiency, memorization mitigation and quality preservation. 
Within the inference-time mitigation category, our method outperforms all prior work. Given this, an important area for future work is to integrate insights from our method into training-time and / or finetuning-time mitigation methods to investigate whether we can push the pareto frontier further for those categories too.

% Our findings are currently limited to diffusion-based T2I models. Future work should assess whether similar trends and trade-offs hold for other types of generative models, including those that do not rely on textual conditioning (e.g., GANs). Additionally, while we adapt two representative unlearning algorithms, further research is needed to explore a wider range of unlearning methods, including those developed for convolutional networks or for higher-level concept unlearning. 

\subsection{Implementation Details} \label{sec:implementation_details}
% \textbf{Compute resources.} 
\textbf{Evaluation configuration and hardware.} 
Following prior work on memorization mitigation \citep{han2025adjusting,ren2024unveiling,wen2024detecting}, we evaluate our methods on two Stable Diffusion (SD) models: SD v1.4 and SD v2.0. 
Since SD v2.0 was pretrained on a de-duplicated dataset, verbatim memorization is substantially reduced compared to SD v1.4; however, it still exhibits template memorization, as noted by \citet{webster2023reproducible}. We therefore evaluate memorization under three settings:
(i) SD v1.4 with verbatim memorization,
(ii) SD v1.4 with template memorization, and
(iii) SD v2.0 with template memorization.

We start from the 500 memorized prompts from the LAION dataset identified by \citet{webster2023reproducible}. 
For each setting, we generate images using the corresponding model architecture and compute SSCD scores for all prompts. To focus on genuinely memorized cases, we retain only prompts with SSCD $>0.7$. This filtering yields 72 prompts for SD v1.4 with verbatim memorization, 143 prompts for SD v1.4 with template memorization, and 96 prompts for SD v2.0 with template memorization.
For all experiments, we generate 4 images per prompt using 50 inference steps. Results are reported as mean values with 95\% confidence intervals for SSCD and CLIP. FID is computed over the entire generation set and therefore does not include confidence intervals.

% \eleni{this is describing the eval right?} 
All experiments were performed on NVIDIA A5000 GPUs, with a total computational cost of approximately 1,500 GPU hours.

\textbf{Batched forward pass.} We concatenate the unconditional, memorized conditional, and spike-attenuated conditional inputs into a single batch, and then enable the U-Net to process all streams simultaneously in one forward pass. 
Within each cross-attention layer, we intercept the intermediate attention logit tensor 
% (which aggregates the attention maps for all streams)
and dynamically slice it along the batch dimension to isolate the spike-attenuated conditional stream. We then compute token-wise statistics (i.e., Z-scores) on this specific slice to detect outlier ``spikes'' and apply an in-place attenuation to the logits immediately before the softmax operation. Thus, we effectively suppress CA of memorization triggers in real-time without disrupting the parallel computation of other streams.
A diagram of the overall inference procedure of CA-in-GUARD is shown in Figure \ref{fig:guard_diagram}.

\textbf{Hyperparameter Tuning.} 
We tune hyperparameters separately for each architecture-memorization setting and each mitigation method.
For CA attenuation, we consider two hyperparameters: the scaling factor $\alpha$, which controls the strength of CA attenuation, and the threshold $\tau$ used for spike detection (see Section \ref{sec:detect-and-attenuate}).
For CA-in-GUARD, in addition to $\alpha$ and $\tau$, we introduce an extra hyperparameter $r$, which controls the strength of the negative target term in the GUARD framework (see Eq. \ref{eq:guard_compose_orig} in Section \ref{sec:ar}).
We perform grid search over method-specific hyperparameter ranges for each setting. The ranges explored for CA attenuation and CA-in-GUARD are summarized in Table \ref{tab:hyperparam_ranges}.

\begin{table}[t]
\centering
\caption{Hyperparameter search ranges for CA attenuation and CA-in-GUARD across different settings.}
\label{tab:hyperparam_ranges}
\footnotesize

\begin{subtable}[t]{\columnwidth}
\centering
\begin{tabular}{lccc}
\toprule
 & $\tau$ & $\alpha$ & $r$ \\
\midrule
SD v1.4 -- verbatim mem.  & $[0.01, 0.05]$ & $[0.05, 0.5]$ & / \\
SD v1.4 -- template mem.  & $[0.5, 3.0]$   & $[0.1, 0.7]$  & / \\
SD v2.0 -- template mem.  & $[0.5, 3.0]$   & $[0.1, 0.7]$  & / \\
\bottomrule
\end{tabular}
\caption{CA attenuation}
\end{subtable}

\vspace{0.6em}

\begin{subtable}[t]{\columnwidth}
\centering
\begin{tabular}{lccc}
\toprule
 & $\tau$ & $\alpha$ & $r$ \\
\midrule
SD v1.4 -- verbatim mem.  & $[0.1, 0.3]$ & $[0.1, 0.5]$ & $[0.1, 2.0]$ \\
SD v1.4 -- template mem.  & $[0.5, 1.5]$ & $[0.5, 0.9]$ & $[0.1, 2.0]$ \\
SD v2.0 -- template mem.  & $[1.5, 3.0]$ & $[0.5, 0.9]$ & $[2.0, 6.0]$ \\
\bottomrule
\end{tabular}
\caption{CA-in-GUARD}
\end{subtable}

\end{table}

% \textbf{Code Availability.} 
\textbf{Code.} The code for reproducing the results is available at:
% \url{https://anonymous.4open.science/r/unlearning_t2i-4A24}
\url{https://github.com/kairanzhao/GUARD}

\subsection{Robustness of CA Attenuation on Non-Memorized Prompts}
An important practical question is how memorization mitigation affects prompts that are not memorized by the model. While prior work typically assumes that memorized examples are identified beforehand and mitigation is applied selectively at inference time, we investigate whether CA attenuation negatively impacts generation quality on non-memorized inputs.

We evaluate this using a subset of 200 prompts from the LAION dataset \citep{schuhmann2022laion}, which was used during pretraining of both SD v1.4 and v2.0. These prompts serve as our non-memorized examples. 
As a reference, we first generate images using the pretrained models without any mitigation. The resulting SSCD scores are very low for both models (0.074 for SD v1.4 and 0.069 for SD v2.0, as shown in Table \ref{tab:non-memorized}), confirming that these prompts are indeed not memorized.

We then apply our CA attenuation method using the same hyperparameters that achieve strong mitigation under the memorization settings, without any tuning for this non-memorized regime. We evaluate performance on the same set of prompts for both SD v1.4 and SD v2.0. The results are reported in Table \ref{tab:non-memorized}. 

\begin{table}[htb]
\centering
\caption{Comparison of no mitigation and CA attenuation on non-memorized prompts for Stable Diffusion v1.4 and v2.0. Results are reported as mean values with 95\% confidence intervals.}
\label{tab:non-memorized}
\resizebox{\columnwidth}{!}{%
\begin{tabular}{l|ccc|ccc}
\toprule
 & \multicolumn{3}{c|}{SD v1.4} & \multicolumn{3}{c}{SD v2.0} \\
Method & SSCD & CLIP & FID & SSCD & CLIP & FID \\
\midrule
No mitigation 
& $0.071 \,{\scriptstyle \pm 0.006}$ 
& $0.299 \,{\scriptstyle \pm 0.010}$ 
& 141.947 
& $0.074 \,{\scriptstyle \pm 0.006}$ 
& $0.322 \,{\scriptstyle \pm 0.006}$ 
& 141.350 \\
CA attenuation 
& $0.069 \,{\scriptstyle \pm 0.006}$ 
& $0.298 \,{\scriptstyle \pm 0.006}$ 
& 142.898 
& $0.072 \,{\scriptstyle \pm 0.006}$ 
& $0.320 \,{\scriptstyle \pm 0.005}$ 
& 139.417 \\

% No mitigation 
% & $0.071 \,{\scriptstyle \pm 0.006}$ & $0.299 \pm 0.010$ & 141.947 
% & $0.074 \pm 0.006$ & $0.322 \pm 0.006$ & 141.350 \\
% CA attenuation 
% & $0.069 \pm 0.006$ & $0.297 \pm 0.006$ & 142.898 
% & $0.072 \pm 0.006$ & $0.320 \pm 0.005$ & 139.417 \\
\bottomrule
\end{tabular}
}
\end{table}

Across all metrics, we observe no statistically significant difference between the no-mitigation baseline and CA attenuation on the non-memorized dataset.
These promising findings suggest that CA attenuation is robust to non-memorized inputs and does not degrade generation quality when memorization is absent. 
% \peter{
\textit{
Importantly, this relaxes a common assumption made by prior work: mitigation does not require (methods to accumulate) prior knowledge of which prompts are memorized. }
% }
Instead, CA attenuation can be applied universally at inference time, providing protection against memorization without harming non-memorized generations.
A systematic evaluation of the potential negative impact of other memorization mitigation methods on non-memorized examples is equally important, but we leave such a comprehensive evaluation to future work.

% \subsection{Analysis of Positive Target Choice} \todo{Other GUARDs, given by e.g paraphrasing or other choices for the positive token.}
% \peter{move this subsection to appendix..?}

\subsection{Alternative Attract Components in GUARD}\label{app:guard-alternative-target}
%To better understand the sources of these gains, w
We next analyze the role of the \textit{attract} component in GUARD through ablations, by using alternative definitions of ``positive targets'' to attract toward.
% as well as controlled studies of cross-attention attenuation outside the AR framework.
We focus on a comparison between two classes of positive targets:
(i) CA-attenuation-based targets, i.e.,  attenuating cross-attention at memorization-relevant token locations, and
(ii) semantic targets, i.e., using a paraphrased sequence of the original memorized prompt and use its predicted noise as the target.
Figure \ref{fig:ca-vs-semantic} compares these two variants across all memorization settings.

\begin{figure*}[t]
\centering
\begin{subfigure}[b]{0.34\textwidth}
 \centering
\includegraphics[scale=0.3]{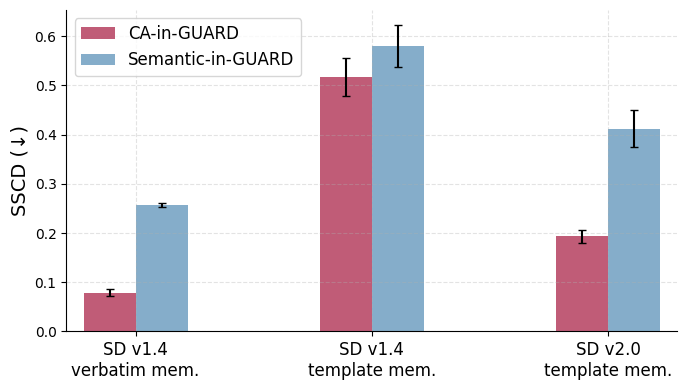}
 \caption{Best SSCD}
 % \label{fig:ar-ca-eccv-sscd}
\end{subfigure}
%  \hfill
\begin{subfigure}[b]{0.34\textwidth}
\centering
\includegraphics[scale=0.3]{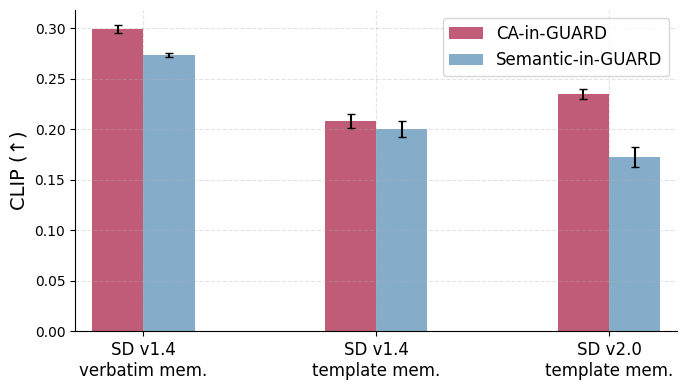}
\caption{Best CLIP}
% \label{fig:ar-ca-eccv-clip}
\end{subfigure}
%  \hfill
\begin{subfigure}[b]{0.3\textwidth}
\centering
\includegraphics[scale=0.3]{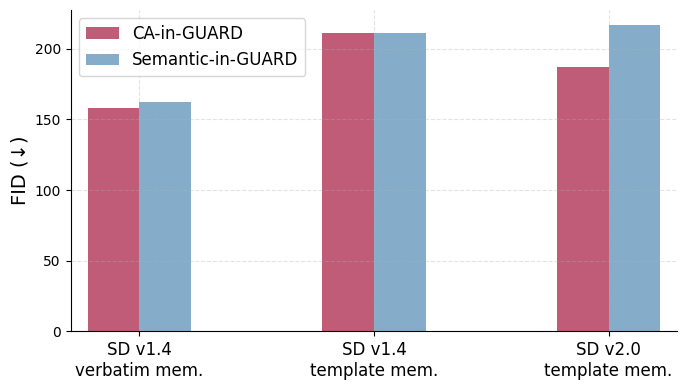}
\caption{Best FID}
% \label{fig:ar-cv-eccv-fid}
\end{subfigure}
\caption{The \textit{best achievable} SSCD, CLIP, and FID of CA-in-GUARD (our default) versus semantic-in-GUARD (ablation),
evaluated across 3 settings: SD v1.4 with verbatim memorization, template memorization, and SD v2.0 with template memorization.}
\label{fig:ca-vs-semantic}
\end{figure*}

As shown in Figure \ref{fig:ca-vs-semantic}, CA-attenuation-based targets consistently outperform semantic targets
as the positive target choice for GUARD across all three settings. 
This suggests that directly manipulating cross-attention dynamics provides more effective memorization mitigation while preserving generation quality, compared to semantic paraphrasing.
In the following, we further ablate different strategies for applying CA attenuation within the GUARD framework.

\subsection{Ablations for CA Attenuation}
\label{app:ca-attenuation-ablations}

\subsubsection{U-Net Block Identification and Ablation}\label{app:block}

\begin{figure*}[t]
\centering
\begin{subfigure}[b]{0.34\textwidth}
 \centering
\includegraphics[scale=0.3]{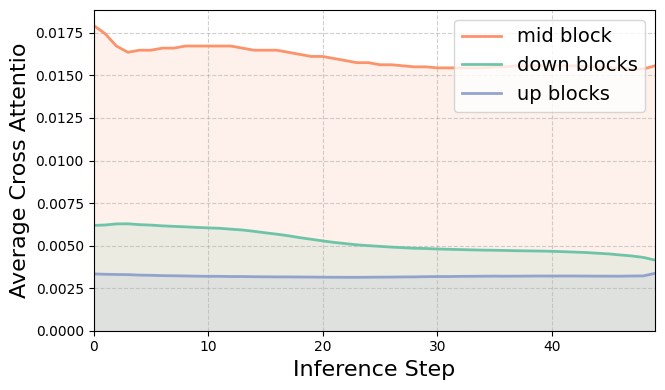}
 \caption{Memorized prompt}
 \label{fig:block-mem}
\end{subfigure}
%  \hfill
\begin{subfigure}[b]{0.34\textwidth}
\centering
\includegraphics[scale=0.3]{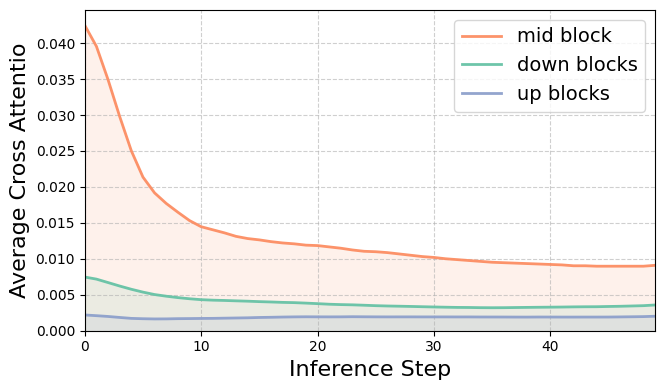}
\caption{Paraphrased prompt}
\label{fig:block-para}
\end{subfigure}
%  \hfill
\begin{subfigure}[b]{0.3\textwidth}
\centering
\includegraphics[scale=0.3]{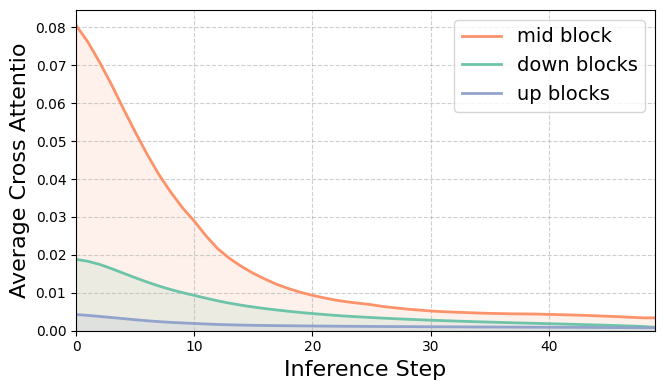}
\caption{Non-memorized prompt}
\label{fig:block-nomem}
\end{subfigure}
\caption{
Cross-attention mass in different U-Net blocks over inference steps, by comparing
three type of prompts: (i) a memorized prompt, (ii) a counterfactual non-memorized prompt obtained by paraphrasing the memorized prompt (which already substantially reduces SSCD), and (iii) an unrelated non-memorized prompt
}
\label{fig:block}
\end{figure*}

Section \ref{sec:ca_attenuation} shows that memorization is associated with abnormally high CA values that persist throughout the generation process, consistent with prior findings \citep{ren2024unveiling}. Attenuating CA at critical locations can therefore mitigate memorization. However, it remains unclear whether such attenuation should be applied uniformly across all U-Net blocks. In particular, a more surgical strategy--selectively attenuating CA in blocks that are primarily responsible for memorization while preserving others that may be more important for generation quality--may achieve stronger mitigation with less degradation in output quality.

To study this, we analyze where memorization-related CA concentrates across the U-Net. We use the EOT token in Stable Diffusion v1.4 under verbatim memorization as a representative example.
% (additional cases are provided in Figure \todo{add more examples}). 
For each U-Net block, we track the average CA mass assigned to EOT over inference steps. 
We compare three prompts: (i) a memorized prompt, (ii) a counterfactual non-memorized prompt obtained by paraphrasing the memorized prompt (which already substantially reduces SSCD), and (iii) an unrelated non-memorized prompt.

The results in Figure \ref{fig:block} reveal a clear structural distinction. For the memorized prompt, CA at EOT is highest in the mid-block, followed by the down-blocks, and remains elevated throughout the generation process. 
In contrast, for both non-memorized prompts (Figures \ref{fig:block-para} and \ref{fig:block-nomem}), CA in the mid- and down-blocks, although sometimes high at early steps, decays rapidly as generation proceeds. 
CA in the remaining blocks is consistently low for both memorized and non-memorized prompts, indicating limited involvement in memorization dynamics. We therefore do not intervene on these blocks.

Motivated by this observation, our intervention selectively attenuates CA only in the mid- and down-blocks, aiming to simulate the natural decay pattern observed in non-memorized generations by gradually reducing CA at memorization-trigger tokens. Figure \ref{fig:blocks} presents an ablation comparing block-selective attenuation with uniform attenuation across all blocks. Selective attenuation consistently yields a more favorable SSCD-CLIP-FID trade-off, as reflected by improved Pareto frontiers.

\begin{figure*}[htb]
\centering
    \begin{subfigure}[b]{0.49\textwidth}
    \centering
    \includegraphics[scale=0.4]{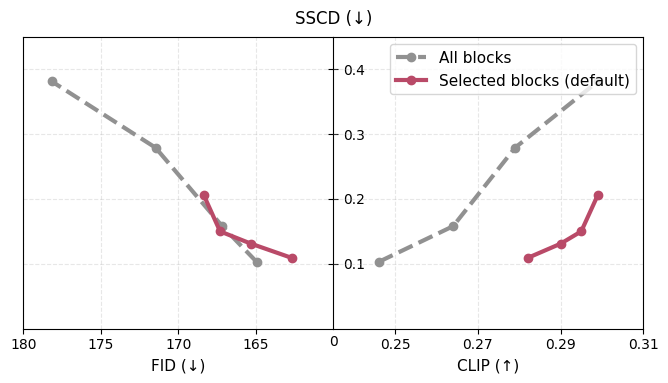}
    \caption{SD v1.4 -- verbatim memorization}
    \label{fig:blocks-verb}
    \end{subfigure}
    \begin{subfigure}[b]{0.49\textwidth}
    \centering
    \includegraphics[scale=0.4]{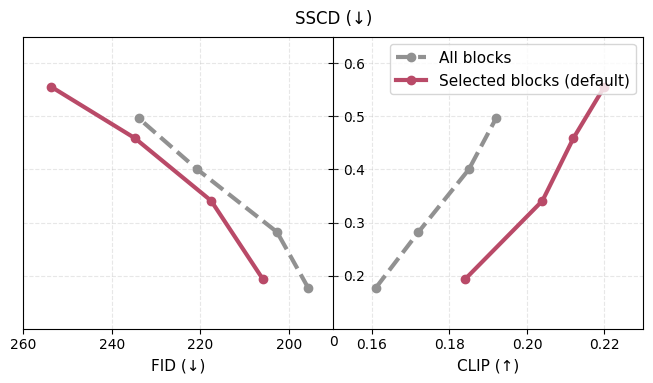}
    \caption{SD v2.0 -- template memorization}
    \label{fig:blocks-temp}
    \end{subfigure}
\caption{\textbf{SSCD–FID and SSCD–CLIP Pareto frontiers comparing CA attenuation using \textit{selective blocks}(our default) vs. \textit{all blocks}(ablation).}
Lower SSCD and FID and higher CLIP indicate better performance; methods closer to the bottom-right corner are optimal in both the left and right subfigures.}
\label{fig:blocks}
\end{figure*}

\subsubsection{Attention Heads Identification and Ablation}\label{app:head}
Following the identification of the most memorization-critical U-Net blocks (mid- and down-blocks) and the corresponding block-level ablations, we next investigate which attention heads within these blocks should be targeted. The goal is to further localize the intervention by focusing on the heads most responsible for memorization, while leaving the remaining heads unaffected.

For each attention head in the selected mid- and down-blocks, we compute a \emph{hotness} score that captures how strongly and selectively the head attends to the memorization trigger token (e.g., the EOT token). Specifically, the score is defined as 
$H_i = \text{AttnScore}_{\text{i}} \times \left(1 / \text{Entropy}\right)$,
where $\text{AttnScore}_{\text{i}}$ denotes the total attention mass assigned to the trigger token $i$, and $\text{Entropy}$ denotes the dispersion of the head’s attention distribution. 
This score prioritizes heads that concentrate attention sharply on the trigger token rather than distributing it diffusely across the prompt.
We then, for each block, select the top $k$ heads with the highest hotness scores and apply CA attenuation only to this subset of ``hot'' heads for the remainder of the generation process. 
We experiment with different proportions $k$ of selected heads per block to study the trade-off between memorization mitigation and generation quality. 

\begin{figure*}[htb]
\centering
    \begin{subfigure}[b]{0.49\textwidth}
    \centering
    \includegraphics[scale=0.4]{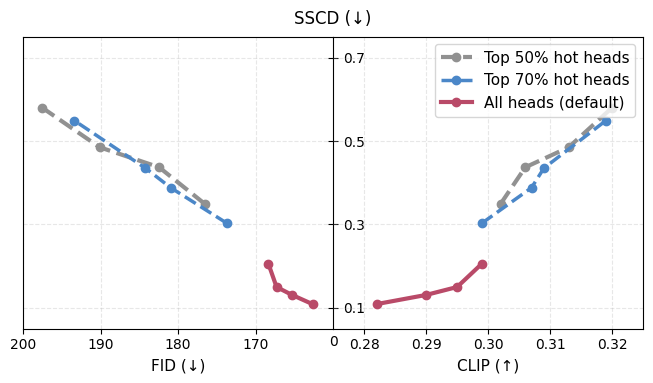}
    \caption{SD v1.4 -- verbatim memorization}
    \label{fig:heads-verb}
    \end{subfigure}
    \begin{subfigure}[b]{0.49\textwidth}
    \centering
    \includegraphics[scale=0.4]{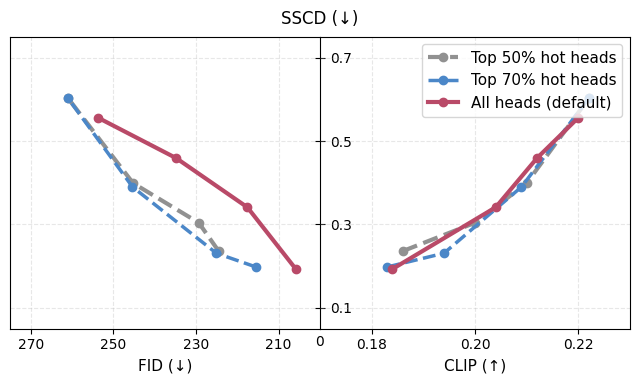}
    \caption{SD v2.0 -- template memorization}
    \label{fig:heads-temp}
    \end{subfigure}
\caption{\textbf{SSCD-FID and SSCD-CLIP Pareto frontiers comparing CA attenuation over \textit{all heads} (our default) vs. \textit{top k\% heads} (ablation)}. 
Lower SSCD and FID and higher CLIP indicate better performance; thus, methods closer to the bottom-right corner are optimal for both left and right subfigures.
}
\label{fig:heads}
\end{figure*}

The results in Figure \ref{fig:heads}.
reveal a clear difference between memorization types. 
Under verbatim memorization, restricting CA attenuation to only a subset of heads substantially limits effectiveness: the lowest SSCD achievable is approximately 0.3, compared to around 0.1 when all heads are attenuated. 
In contrast, under template memorization, performance is much less sensitive to the number of selected heads. 
These observations indicate that selectively targeting heads is insufficient for robust mitigation in the verbatim memorization setting, and that attenuating all heads within the identified blocks is a stronger and more reliable default.

\subsubsection{Timesteps Identification and Ablation}\label{app:step}
We next examine how the effectiveness of CA attenuation varies depending on when it is applied during the diffusion inference process. Specifically, we consider four strategies:
(i) applying CA attenuation at all timesteps (our default);
(ii) applying CA attenuation only during the first 50\% of timesteps;
(iii) applying CA attenuation at all timesteps with a cosine decay schedule; and
(iv) applying CA attenuation at all timesteps with a linear decay schedule.

Figure \ref{fig:timesteps} shows the corresponding SSCD–CLIP and SSCD–FID Pareto frontiers. Overall, applying CA attenuation at all timesteps emerges as a strong and robust default, as it consistently yields more favorable SSCD-CLIP trade-offs compared to other strategies.

\begin{figure*}[htb]
\centering
    \begin{subfigure}[b]{0.49\textwidth}
    \centering
    \includegraphics[scale=0.4]{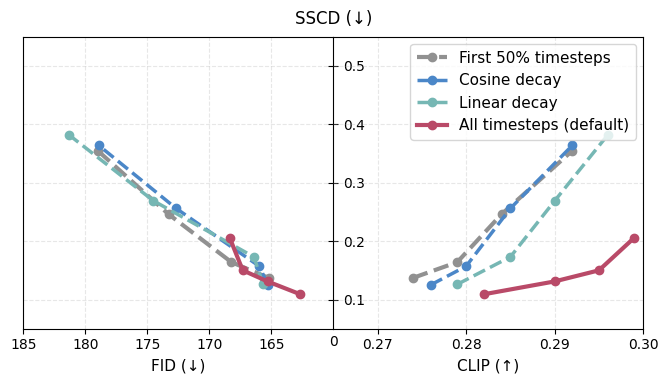}
    \caption{SD v1.4 -- verbatim memorization}
    \label{fig:ts-verb}
    \end{subfigure}
    \begin{subfigure}[b]{0.49\textwidth}
    \centering
    \includegraphics[scale=0.4]{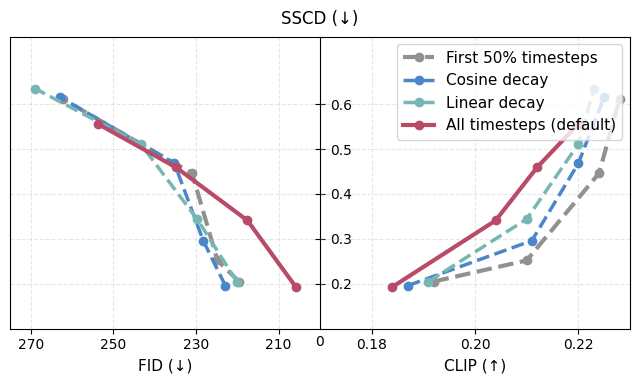}
    \caption{SD v2.0 -- template memorization}
    \label{fig:ts-temp}
    \end{subfigure}
\caption{\textbf{SSCD-FID and SSCD-CLIP Pareto frontiers comparing CA attenuation over \textit{all timesteps} (our default) vs. \textit{other variants} (ablation)}. 
Lower SSCD and FID and higher CLIP indicate better performance; thus, methods closer to the bottom-right corner are optimal for both left and right subfigures.
}
\label{fig:timesteps}
\end{figure*}

% \subsection{Ablation Study: GUARD vs negative term only}

% \kairan{
\subsection{Ablations for CA-in-GUARD}
\label{app:ca-in-guard-ablations}

For comparability with prior work, our main experiments use the standard inference configuration: DDIM sampling with 50 denoising steps and classifier-free guidance (CFG) scale $s=7.5$. 
To evaluate whether CA-in-GUARD depends on this specific setup, we conduct additional ablations over common Stable Diffusion inference choices, including samplers, denoising step counts, guidance schedules, and CFG scales. Results are shown in Figure \ref{fig:guard-ablation}. 
Overall, CA-in-GUARD remains robust and performs well across these settings.

\textbf{Sampler.}
We compare DDIM, Euler A, and DPM++ samplers. CA-in-GUARD consistently reduces memorization across these samplers, indicating that its effectiveness is not tied to a particular sampling algorithm.

\textbf{Number of denoising steps.}
We evaluate 30, 50, and 70 denoising steps. 
CA-in-GUARD remains effective under both shorter and longer sampling budgets, suggesting that the method is stable across different inference costs.

\textbf{Guidance schedule.}
We compare three guidance schedules: no schedule, cosine decay, and linear decay. 
CA-in-GUARD performs reliably across these schedules, showing that the method does not require a finely tuned guidance trajectory.

\textbf{CFG scale.}
We further ablate CFG scales $s \in \{1,4,7,7.5\}$, where $s=7.5$ is the default setting. Without mitigation, reducing $s$ lowers memorization but also degrades generation quality. This behavior is expected from the standard CFG update,
\[
\epsilon_{\mathrm{pred}}
=
\epsilon_{\mathrm{uncond}}
+
s(\epsilon_{\mathrm{cond}}-\epsilon_{\mathrm{uncond}}),
\]
where $s$ controls the strength of prompt-conditioned steering. Smaller values of $s$ weaken prompt alignment, which can reduce memorization but also harms fidelity.

CA-in-GUARD remains effective across CFG scales, consistently reducing SSCD relative to the no-mitigation baseline while maintaining a favorable quality trade-off. Its mitigation effect is weaker at $s=1$, but this behavior is not specific to CA-in-GUARD: other inference-time mitigation methods also show limited improvement over the ``no mitigation'' baseline (Table \ref{tab:cfg-scale-ablation-s1}). 
We interpret this as a consequence of the reduced prompt-conditioned guidance signal. 
Larger CFG scales amplify the prompt-conditioned direction, increasing both memorization risk and the structured signal that inference-time mitigation methods can intervene on. 
At $s=1$, CFG reduces to plain conditional denoising, leaving less headroom for any inference-time method to operate on (because most inference-time methods operate by modifying the guidance signal or its induced trajectory). We leave a more systematic study of this interaction as future work.

\begin{figure*}[htb]
\centering
    \begin{subfigure}[b]{0.49\textwidth}
    \centering
    \includegraphics[scale=0.4]{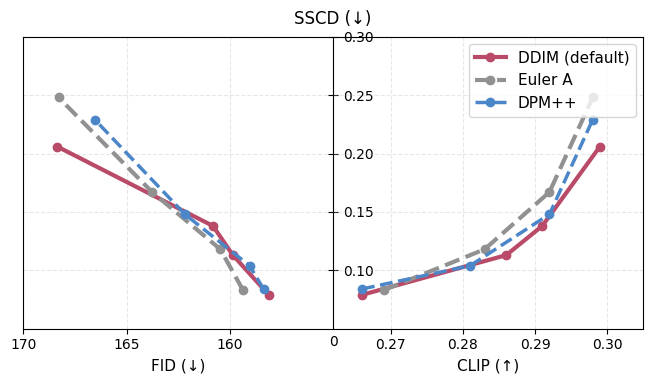}
    \caption{Sampler Ablation}
    % \label{fig:ts-verb}
    \end{subfigure}
    \begin{subfigure}[b]{0.49\textwidth}
    \centering
    \includegraphics[scale=0.4]{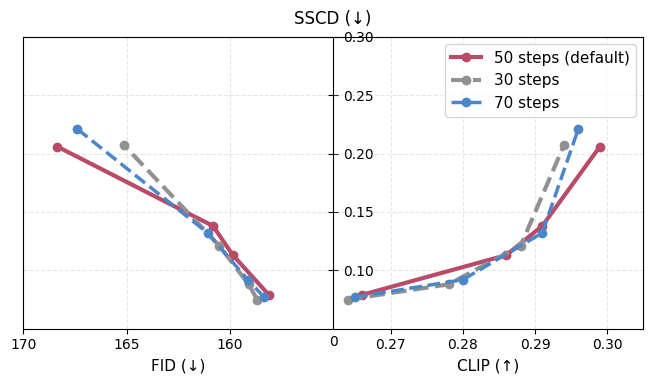}
    \caption{Inference Step Count Ablation}
    % \label{fig:ts-temp}
    \end{subfigure}
    \begin{subfigure}[b]{0.49\textwidth}
    \centering
    \includegraphics[scale=0.4]{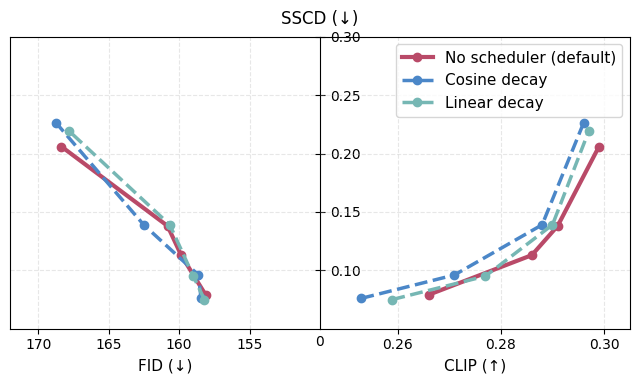}
    \caption{Scheduler Ablation}
    % \label{fig:ts-temp}
    \end{subfigure}
    \begin{subfigure}[b]{0.49\textwidth}
    \centering
    \includegraphics[scale=0.4]{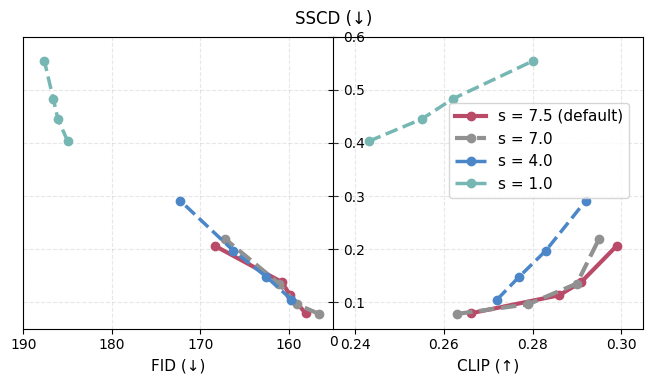}
    \caption{CFG Scale Ablation}
    \label{fig:cfg-scale-ablation}
    \end{subfigure}
\caption{SSCD-FID and SSCD-CLIP Pareto frontiers for CA-in-GUARD ablations across \textit{sampler choice}, \textit{inference steps}, \textit{scheduler choice}, and \textit{CFG scale}. Lower SSCD and FID, together with higher CLIP, indicate better performance; methods closer to the bottom-right corner achieve a more favorable trade-off.}
\label{fig:guard-ablation}
\end{figure*}

\begin{table}[htb]
\centering
\caption{Quantitative results for the CFG scale ablation in Figure \ref{fig:cfg-scale-ablation}, reported at CFG scale $s=1.0$, with each method shown under the configuration with the best SSCD (our primary memorization-mitigation metric).}
\label{tab:cfg-scale-ablation-s1}
\footnotesize
\resizebox{\columnwidth}{!}{%
\begin{tabular}{l|ccc}
\toprule
Method & SSCD & CLIP & FID \\
\midrule
No mitigation & $0.571 \pm 0.068$ & $0.280 \pm 0.019$ & $194.697$ \\
Wen et al.    & $0.569 \pm 0.064$ & $0.284 \pm 0.016$ & $188.374$ \\
CA-in-GUARD   & $0.403 \pm 0.071$ & $0.243 \pm 0.019$ & $184.982$ \\
\bottomrule
\end{tabular}
}
\end{table}
% }

\subsection{Detailed Results}

\begin{table*}[htb]
\centering
\caption{Comparison of memorization mitigation methods across SD v1.4 and SD v2.0 under verbatim and template memorization settings. Results are evaluated using SSCD, CLIP, and FID. Each subtable shows results when selecting the best configuration for one target metric (SSCD, CLIP, or FID). For each prompt, we generate four images and report the mean $\pm$
% $\ci{\cdot}{\cdot}$ 
(95\% confidence interval) across the four generations.}
\label{tab:itmm}

% =========================
% Best SSCD
% =========================
\begin{subtable}[t]{\textwidth}
\centering
\footnotesize
\resizebox{\textwidth}{!}{%
\begin{tabular}{l|ccc|ccc|ccc}
\toprule
 & \multicolumn{3}{c|}{SD v1.4 -- verbatim memorization}
 & \multicolumn{3}{c|}{SD v1.4 -- template memorization}
 & \multicolumn{3}{c}{SD v2.0 -- template memorization} \\
Method
 & SSCD ($\downarrow$) & CLIP ($\uparrow$) & FID ($\downarrow$)
 & SSCD ($\downarrow$) & CLIP ($\uparrow$) & FID ($\downarrow$)
 & SSCD ($\downarrow$) & CLIP ($\uparrow$) & FID ($\downarrow$)\\
\midrule

No mitigation
 & \ci{0.875}{0.001} & \ci{0.346}{0.001} & 243.056
 & \ci{0.776}{0.017} & \ci{0.219}{0.007} & 258.976
 & \ci{0.735}{0.011} & \ci{0.215}{0.005} & 303.266 \\

RTA
 & \ci{0.328}{0.007} & \ci{0.263}{0.002} & 175.866
 & \ci{0.617}{0.043} & \ci{0.187}{0.010} & 218.343
 & \ci{0.543}{0.048} & \ci{0.183}{0.009} & 233.580 \\

Wen et al.
 & \ci{0.115}{0.011} & \ci{0.267}{0.003} & 162.848
 & \ci{0.545}{0.038} & \ci{0.188}{0.008} & 209.719
 & \ci{0.260}{0.026} & \ci{0.183}{0.008} & 188.914 \\

Ren et al.
 & \ci{0.113}{0.007} & \ci{0.258}{0.005} & 164.638
 & \ci{0.602}{0.033} & \ci{0.184}{0.007} & 222.066
 & \ci{0.356}{0.024} & \ci{0.188}{0.007} & 208.416 \\

Han et al.
 & \ci{0.191}{0.016} & \ci{0.256}{0.008} & 166.551
 & \ci{0.479}{0.033} & \ci{0.188}{0.006} & 210.839
 & \ci{0.401}{0.024} & \ci{0.186}{0.005} & 208.852 \\

\midrule

CA attenuation
 & \ci{0.109}{0.006} & \ci{0.282}{0.004} & 164.660
 & \ci{0.530}{0.038} & \ci{0.185}{0.009} & 212.240
 & \ci{0.193}{0.014} & \ci{0.184}{0.005} & 245.850 \\

CA-in-GUARD
 & \ci{0.079}{0.007} & \ci{0.266}{0.015} & 158.115
 & \ci{0.517}{0.038} & \ci{0.186}{0.008} & 210.983
 % & \ci{0.223}{0.017} & \ci{0.183}{0.006} & 187.250
  % & \ci{0.193}{0.014} & \ci{0.183}{0.005} & 212.727
 & \ci{0.193}{0.014} & \ci{0.183}{0.005} & 212.727
 \\

\bottomrule
\end{tabular}
}
\vspace{0.5pt}
\caption{Best SSCD 
% \kairan{just a note: for the updates in red in this table, i selected another data point with similar SSCD and CLIP but much better FID compared to CA attenuation, so the CA and GUARD here are from different runs}
}
\label{tab:itmm-sscd}
\end{subtable}

\vspace{3pt}

% =========================
% Best CLIP
% =========================
\begin{subtable}[t]{\textwidth}
\centering
\footnotesize
\resizebox{\textwidth}{!}{%
\begin{tabular}{l|ccc|ccc|ccc}
\toprule
 & \multicolumn{3}{c|}{SD v1.4 -- verbatim memorization}
 & \multicolumn{3}{c|}{SD v1.4 -- template memorization}
 & \multicolumn{3}{c}{SD v2.0 -- template memorization} \\
Method
 & SSCD ($\downarrow$) & CLIP ($\uparrow$) & FID ($\downarrow$)
 & SSCD ($\downarrow$) & CLIP ($\uparrow$) & FID ($\downarrow$)
 & SSCD ($\downarrow$) & CLIP ($\uparrow$) & FID ($\downarrow$)\\
\midrule

No mitigation
 & \ci{0.875}{0.001} & \ci{0.346}{0.001} & 243.056
 & \ci{0.776}{0.017} & \ci{0.219}{0.007} & 258.976
 & \ci{0.735}{0.011} & \ci{0.215}{0.005} & 303.266 \\

RTA
 & \ci{0.526}{0.003} & \ci{0.290}{0.001} & 196.322
 & \ci{0.699}{0.033} & \ci{0.204}{0.009} & 237.061
 & \ci{0.660}{0.034} & \ci{0.205}{0.005} & 269.205 \\

Wen et al.
 & \ci{0.333}{0.022} & \ci{0.299}{0.003} & 174.106
 & \ci{0.661}{0.032} & \ci{0.207}{0.007} & 230.140
 & \ci{0.630}{0.024} & \ci{0.218}{0.004} & 263.483 \\

Ren et al.
 & \ci{0.337}{0.007} & \ci{0.301}{0.002} & 172.417
 & \ci{0.655}{0.033} & \ci{0.194}{0.007} & 232.503
 & \ci{0.584}{0.024} & \ci{0.204}{0.005} & 254.504 \\

Han et al.
 & \ci{0.377}{0.014} & \ci{0.298}{0.004} & 182.813
 & \ci{0.662}{0.024} & \ci{0.209}{0.006} & 235.635
 & \ci{0.656}{0.018} & \ci{0.209}{0.005} & 273.532 \\

\midrule

CA attenuation
 & \ci{0.206}{0.004} & \ci{0.299}{0.004} & 168.356
 & \ci{0.679
 }{0.025} & \ci{0.206}{0.008} & 239.889
 & \ci{0.556}{0.023} & \ci{0.220}{0.005} & 253.737 \\

CA-in-GUARD
 % & \ci{0.220}{0.004} & \ci{0.299}{0.002} & 166.904
 & \ci{0.206}{0.004} & \ci{0.299}{0.004} & 168.356
 & \ci{0.667}{0.029} & \ci{0.208}{0.007} & 236.523
 & \ci{0.558}{0.028} & \ci{0.235}{0.005} & 243.897 \\

\bottomrule
\end{tabular}
}
\vspace{0.5pt}
\caption{Best CLIP}
\label{tab:itmm-clip}
\end{subtable}

\vspace{3pt}

% =========================
% Best FID
% =========================
\begin{subtable}[t]{\textwidth}
\centering
\footnotesize
\resizebox{\textwidth}{!}{%
\begin{tabular}{l|ccc|ccc|ccc}
\toprule
 & \multicolumn{3}{c|}{SD v1.4 -- verbatim memorization}
 & \multicolumn{3}{c|}{SD v1.4 -- template memorization}
 & \multicolumn{3}{c}{SD v2.0 -- template memorization} \\
Method
 & SSCD ($\downarrow$) & CLIP ($\uparrow$) & FID ($\downarrow$)
 & SSCD ($\downarrow$) & CLIP ($\uparrow$) & FID ($\downarrow$)
 & SSCD ($\downarrow$) & CLIP ($\uparrow$) & FID ($\downarrow$) \\
\midrule

No mitigation
 & \ci{0.875}{0.001} & \ci{0.346}{0.001} & 243.056
 & \ci{0.776}{0.017} & \ci{0.219}{0.007} & 258.976
 & \ci{0.735}{0.011} & \ci{0.215}{0.005} & 303.266 \\

RTA
 & \ci{0.328}{0.007} & \ci{0.263}{0.002} & 175.866
 & \ci{0.617}{0.043} & \ci{0.187}{0.010} & 218.343
 & \ci{0.543}{0.048} & \ci{0.183}{0.009} & 233.580 \\

Wen et al.
 & \ci{0.115}{0.011} & \ci{0.267}{0.003} & 162.848
 & \ci{0.545}{0.038} & \ci{0.188}{0.008} & 209.719
 & \ci{0.292}{0.027} & \ci{0.191}{0.007} & 197.186 \\

Ren et al.
 & \ci{0.142}{0.003} & \ci{0.266}{0.004} & 163.670
 & \ci{0.602}{0.033} & \ci{0.184}{0.007} & 222.066
 & \ci{0.356}{0.024} & \ci{0.188}{0.007} & 208.416 \\

Han et al.
 & \ci{0.191}{0.016} & \ci{0.256}{0.008} & 166.551
 & \ci{0.479}{0.033} & \ci{0.188}{0.006} & 210.839
 & \ci{0.401}{0.024} & \ci{0.186}{0.005} & 208.852 \\

\midrule

CA attenuation
 & \ci{0.110}{0.007} & \ci{0.284}{0.001} & 162.825
 & \ci{0.530}{0.038} & \ci{0.185}{0.009} & 212.240
 & \ci{0.341}{0.021} & \ci{0.204}{0.005} & 217.558 \\

CA-in-GUARD
 & \ci{0.079}{0.007} & \ci{0.266}{0.015} & 158.115
 & \ci{0.517}{0.038} & \ci{0.186}{0.008} & 210.983
 & \ci{0.223}{0.017} & \ci{0.183}{0.006} & 187.250 \\

\bottomrule
\end{tabular}
}
\vspace{0.5pt}
\caption{Best FID
% \peter{we should somehow mention that the config that mins FID often tends to be the one that mins SSCD as well for most algos...else astute reviewers may think something is off here }
}
\label{tab:itmm-fid}
\end{subtable}
\label{tab:main_results}
\end{table*}

\subsubsection{CA-in-GUARD vs. CA Attenuation vs. Prior Work}
\label{app:ar-ca-eccv}

This section presents Figure \ref{fig:ar-ca-eccv}, a focused subset of the results from Figure \ref{fig:best_achievable}, restricted to the methods most relevant for addressing the two specific questions in Section \ref{sec:results}:
(1) whether CA attenuation alone improves upon prior work, and
(2) whether further gains can be achieved by incorporating CA attenuation into the GUARD framework (CA-in-GUARD).

To address the first question, we compare CA attenuation with \citet{ren2024unveiling}, which is the closest prior approach in spirit. 
As shown in Figure \ref{fig:ar-ca-eccv}, CA attenuation consistently matches or outperforms \citet{ren2024unveiling} across all architecture versions and memorization settings. 
The improvement is particularly pronounced in memorization mitigation (Figure \ref{fig:ar-ca-eccv-sscd}), where CA attenuation yields substantially lower SSCD under template memorization for both SD v1.4 and SD v2.0.

We then compare CA attenuation with CA-in-GUARD. Note that CA attenuation corresponds to a special case of CA-in-GUARD obtained by setting $r = 0$ in Eq. \ref{eq:guard_compose_orig} (i.e., removing the negative target from GUARD). 
% Therefore, CA attenuation can be viewed as a corner case of CA-in-GUARD. 
As shown in Figure \ref{fig:ar-ca-eccv}, CA-in-GUARD consistently improves upon or maintains the performance of CA attenuation across all metrics and settings, with particularly clear gains in CLIP and FID under SD v2.0.

\begin{figure*}[t]
\centering
\begin{subfigure}[b]{0.34\textwidth}
 \centering
\includegraphics[scale=0.3]{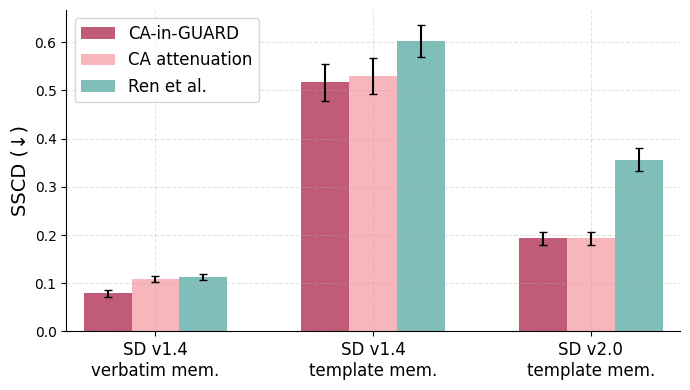}
 \caption{Best SSCD }
 \label{fig:ar-ca-eccv-sscd}
\end{subfigure}
%  \hfill
\begin{subfigure}[b]{0.34\textwidth}
\centering
\includegraphics[scale=0.3]{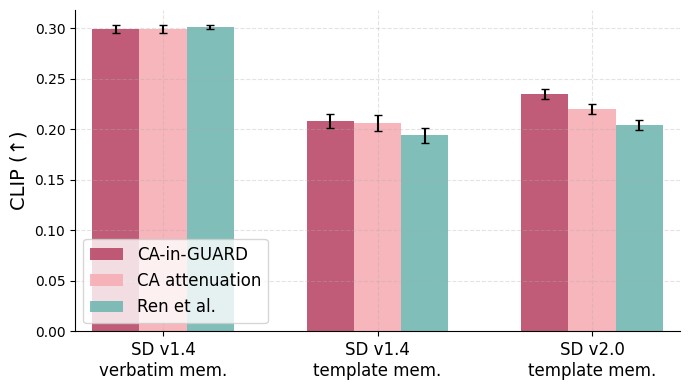}
\caption{Best CLIP }
\label{fig:ar-ca-eccv-clip}
\end{subfigure}
%  \hfill
\begin{subfigure}[b]{0.3\textwidth}
\centering
\includegraphics[scale=0.3]{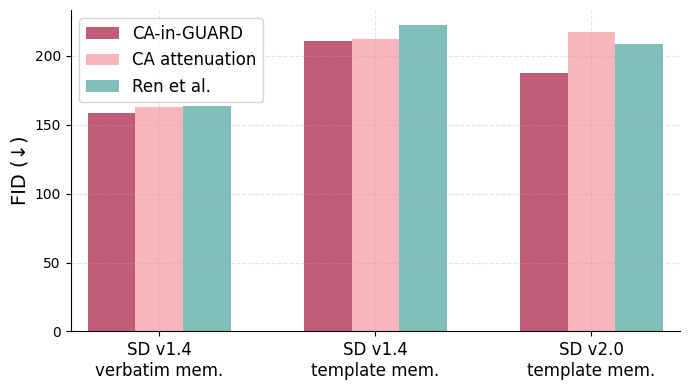}
\caption{Best FID }
\label{fig:ar-cv-eccv-fid}
\end{subfigure}
\caption{\textbf{The \textit{best achievable} SSCD, CLIP, and FID of our methods compared to the prior state-of-the-art (and most related) baseline of Ren et al.}. We plot the best achievable value on each metric \textit{individually}, using the configuration that yields best results on that metric.
Because in each subplot, a different configuration may be used for a given method (the one that yields the best SSCD, best CLIP and best FID, respectively), this plot does not speak to the ability to do well on all metrics \textit{jointly}, nor to trade-offs between these metrics, which we investigate later.}
\label{fig:ar-ca-eccv}
\end{figure*}

\subsubsection{Evaluation via SSCD-CLIP and SSCD-FID Pareto Frontiers}
\label{app:all-methods}

We present here the full results Table \ref{tab:main_results} obtained under our experimental protocol described in Section \ref{sec:protocol}. This table provides detailed numerical results and serves as the basis for the summarized comparisons shown in Figures \ref{fig:best_achievable}, \ref{fig:standalone_example} and \ref{fig:ar-ca-eccv}.

In addition to our protocol-driven evaluation, we also report results using the traditional Pareto-front analysis commonly adopted in prior work \citep{wen2024detecting,ren2024unveiling,han2025adjusting}. 
Specifically, we construct SSCD-CLIP and SSCD-FID Pareto frontiers for each model version and memorization type setting. These results are shown in Figure \ref{fig:pareto-all}.

\begin{figure*}[htb]
\centering
\begin{subfigure}[b]{0.6\textwidth}
\centering
\includegraphics[scale=0.6]{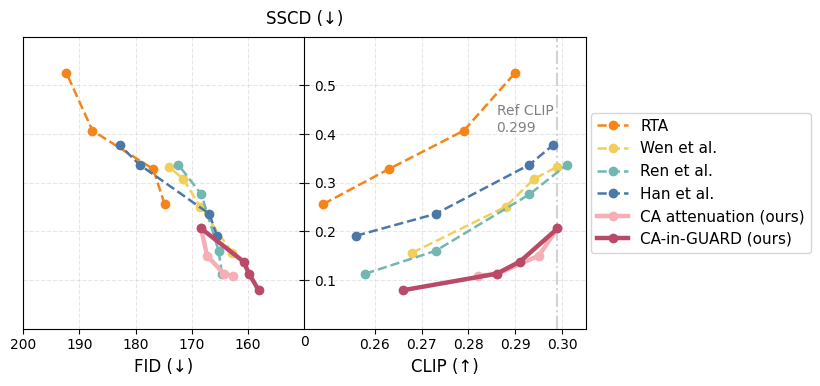}
\caption{SD v1.4 -- verbatim memorization}
% \label{fig:ar-cv-eccv-fid}
\end{subfigure}
%  \hfill
\begin{subfigure}[b]{0.6\textwidth}
\centering
\includegraphics[scale=0.6]{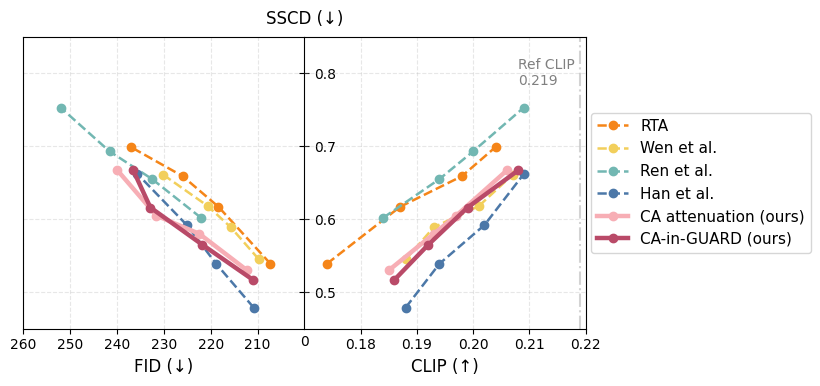}
\caption{SD v1.4 -- template memorization}
% \label{fig:ar-cv-eccv-fid}
\end{subfigure}
%  \hfill
\begin{subfigure}[b]{0.6\textwidth}
\centering
\includegraphics[scale=0.6]{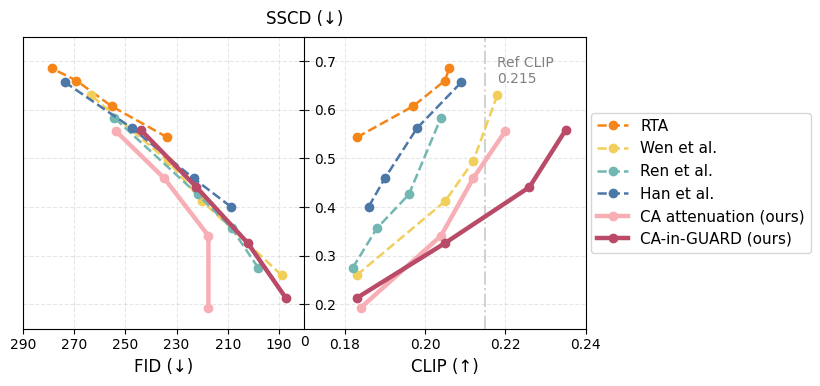}
\caption{SD v2.0 -- template memorization}
% \label{fig:ar-cv-eccv-fid}
\end{subfigure}
\caption{\textbf{SSCD–FID and SSCD–CLIP Pareto frontiers comparing our methods with existing baselines under three memorization settings.}
Lower SSCD and FID and higher CLIP indicate better performance; methods closer to the bottom-right corner represent more favorable trade-offs in both subfigures. 
The \emph{reference CLIP score} indicates the target semantic fidelity that methods are expected to attain (see Section \ref{sec:protocol} -- Hyperparameter selection for details on how this reference is defined).
}
\label{fig:pareto-all}
\end{figure*}

Across all experimental settings, our methods achieve the best or comparable SSCD-CLIP and SSCD-FID trade-offs relative to existing baselines. 
The advantage is particularly notable for the SSCD-CLIP trade-off, where both CA attenuation and CA-in-GUARD consistently dominate other methods. 
The only exception occurs in the SD v1.4 template memorization setting for SSCD–CLIP, where \citet{han2025adjusting} attains a slightly better trade-off but performs significantly worse in most other settings. 
Overall, these results confirm that our methods remain competitive--and often superior--under traditional evaluation protocols, in addition to the more controlled evaluation protocol adopted in Section \ref{sec:protocol}.

% \kairan{
\subsubsection{Evaluation under SSCD and DINO Metrics}
\label{sec:sscd-dino-evaluation}

To further validate the effectiveness of our methods under complementary perceptual similarity metrics, we evaluate template memorization using both SSCD and DINO-based retrieval scores. SSCD is widely used to measure perceptual similarity for verbatim and near-verbatim memorization, while DINO retrieval provides an additional feature-space measure of similarity to training-set neighborhoods \citep{caron2021emerging} . Specifically, we report DINO@1 and DINO@5, corresponding to the retrieval similarity to the nearest and top-5 nearest training examples, respectively. Lower scores indicate weaker similarity to the training set and therefore less memorization.

As shown in Table \ref{tab:sscd-dino}, the conclusions are consistent across both SSCD and DINO metrics. CA attenuation substantially reduces memorization compared with the no-mitigation baseline and remains competitive with prior mitigation methods. CA-in-GUARD further improves or matches this performance across most settings, achieving the strongest overall results for both SD v1.4 and SD v2.0. These results confirm that our methods reduce template memorization not only under SSCD, but also under an independent DINO-based retrieval evaluation.

\begin{table*}[htb]
\centering
\caption{SSCD and DINO results for template memorization settings, as an expansion of Table \ref{tab:main_results_shorter}. Lower values indicate less similarity to the training set and therefore less memorization.}
\label{tab:sscd-dino}
\footnotesize
\resizebox{0.85\textwidth}{!}{%
% \resizebox{\columnwidth}{!}{%
\begin{tabular}{l|ccc|ccc}
\toprule
\multirow{2}{*}{Method}
& \multicolumn{3}{c|}{SD v1.4 – template memorization}
& \multicolumn{3}{c}{SD v2.0 – template memorization} \\
& SSCD ($\downarrow$) & DINO@1 ($\downarrow$) & DINO@5 ($\downarrow$)
& SSCD ($\downarrow$) & DINO@1 ($\downarrow$) & DINO@5 ($\downarrow$)\\
\midrule
No mitigation  & $0.776{\pm}0.017$ & $0.924{\pm}0.006$ & $0.906{\pm}0.013$ & $0.735{\pm}0.011$ & $0.903{\pm}0.006$ & $0.895{\pm}0.007$ \\
RTA            & $0.617{\pm}0.043$ & $0.836{\pm}0.030$ & $0.813{\pm}0.035$ & $0.543{\pm}0.048$ & $0.794{\pm}0.038$ & $0.778{\pm}0.039$ \\
Wen et al.     & $0.545{\pm}0.038$ & $0.823{\pm}0.024$ & $0.790{\pm}0.028$ & $0.260{\pm}0.026$ & $0.655{\pm}0.023$ & $0.637{\pm}0.024$ \\
Ren et al.     & $0.602{\pm}0.033$ & $0.858{\pm}0.015$ & $0.832{\pm}0.020$ & $0.356{\pm}0.024$ & $0.711{\pm}0.020$ & $0.695{\pm}0.020$ \\
Han et al.     & $0.479{\pm}0.033$ & $0.832{\pm}0.015$ & $0.789{\pm}0.019$ & $0.401{\pm}0.024$ & $0.745{\pm}0.017$ & $0.732{\pm}0.018$ \\
\midrule
CA attenuation & $0.530{\pm}0.038$ & $0.811{\pm}0.024$ & $0.774{\pm}0.029$ & $0.193{\pm}0.014$ & $0.647{\pm}0.016$ & $0.626{\pm}0.017$ \\
CA-in-GUARD    & $0.517{\pm}0.038$ & $0.782{\pm}0.023$ & $0.752{\pm}0.026$ & $0.193{\pm}0.014$ & $0.647{\pm}0.016$ & $0.626{\pm}0.017$ \\
\bottomrule
\end{tabular}
}
\end{table*}
% }

% \kairan{
\subsubsection{Full-spectrum Evaluation}
\label{sec:full-spectrum-evaluation}

To examine whether our conclusions hold beyond the high-memorization subset used in the main evaluation in Section \ref{sec:protocol}, we additionally evaluate all 500 examples from \citet{webster2023reproducible}, providing a full-spectrum view across low-, medium-, and high-memorization cases. For each method and model architecture, we select the configuration that achieves the best SSCD score, since SSCD is our primary metric for memorization mitigation.

Table \ref{tab:full-spectrum} reports SSCD, CLIP, and FID results across both SD v1.4 and SD v2.0 architectures. The results are consistent with our main findings: CA attenuation and CA-in-GUARD substantially reduce SSCD compared with the no-mitigation baseline, while maintaining competitive CLIP alignment and FID. In particular, CA-in-GUARD achieves the best SSCD and FID across both architectures, indicating that the proposed method reduces memorization without degrading overall generation quality. These full-spectrum results further support the robustness of our conclusions.

\begin{table*}[htb]
\centering
\caption{Full-spectrum evaluation results across metrics and architectures. For each setting, we select for each method the configuration that yields the best SSCD, since SSCD is the primary metric for memorization mitigation.}
\label{tab:full-spectrum}
\footnotesize
\resizebox{0.85\textwidth}{!}{%
\begin{tabular}{l|ccc|ccc}
\toprule
\multirow{2}{*}{Method}
& \multicolumn{3}{c|}{SD v1.4}
& \multicolumn{3}{c}{SD v2.0} \\
& SSCD ($\downarrow$) & CLIP ($\uparrow$) & FID ($\downarrow$)
& SSCD ($\downarrow$) & CLIP ($\uparrow$) & FID ($\downarrow$)\\
\midrule
No mitigation  & $0.524{\pm}0.028$ & $0.257{\pm}0.006$ & $198.081$ & $0.333{\pm}0.025$ & $0.260{\pm}0.005$ & $146.527$ \\
RTA            & $0.366{\pm}0.026$ & $0.223{\pm}0.007$ & $153.361$ & $0.274{\pm}0.023$ & $0.223{\pm}0.006$ & $135.615$ \\
Wen et al.     & $0.255{\pm}0.019$ & $0.220{\pm}0.006$ & $137.464$ & $0.152{\pm}0.011$ & $0.219{\pm}0.006$ & $126.339$ \\
Ren et al.     & $0.350{\pm}0.024$ & $0.223{\pm}0.006$ & $144.830$ & $0.200{\pm}0.016$ & $0.218{\pm}0.006$ & $130.757$ \\
Han et al.     & $0.239{\pm}0.018$ & $0.220{\pm}0.006$ & $143.297$ & $0.176{\pm}0.023$ & $0.221{\pm}0.006$ & $125.894$ \\
\midrule
CA attenuation & $0.242{\pm}0.018$ & $0.220{\pm}0.007$ & $140.381$ & $0.153{\pm}0.014$ & $0.219{\pm}0.006$ & $128.503$ \\
CA-in-GUARD    & $0.201{\pm}0.014$ & $0.221{\pm}0.007$ & $135.608$ & $0.137{\pm}0.009$ & $0.222{\pm}0.006$ & $125.402$ \\
\bottomrule
\end{tabular}
}
\end{table*}
% }

\subsubsection{Computational efficiency}
\label{app:runtime}

% \textbf{Computational efficiency.} 
% \kairan{
We compare the runtime and memory efficiency of our methods against all baselines, using wall-clock generation time per example and peak GPU memory usage under the default experimental setup described in Section \ref{sec:protocol}. For each method, we measure the inference time and GPU memory consumption across both model architectures, and report the results in Table \ref{tab:itmm-runtime}.
% }

Overall, CA attenuation incurs minimal overhead, as it only manipulates cross-attention on-the-fly during generation. 
CA-in-GUARD is moderately more expensive due to the additional guidance components, 
% but remains computationally efficient in practice. 
but its memory usage remains close to the default generation setting and substantially lower than methods with heavier optimization or adjustment procedures.
Notably, the implementation strategy described in Section \ref{sec:implementation_details} allows the required forward passes to be batched efficiently, which could substantially reduce runtime overhead.

From a wall-clock perspective, the methods can be ranked from fastest to slowest as follows: RTA, CA attenuation, \citet{wen2024detecting}, CA-in-GUARD, \citet{han2025adjusting}, and \citet{ren2024unveiling}. 
In terms of GPU memory usage, CA attenuation is comparable to the no-mitigation setting, while CA-in-GUARD remains lightweight relative to most baselines. 
These results indicate that our approaches achieve strong memorization mitigation while maintaining competitive inference and memory efficiency.
% }

% \peter{Kairan, please describe the setup, what's included in these times exactly, this is the toal time for how many prompts and how many runs per prompt, etc.}
% \peter{conclusions are that our implementation trick to batch the 3 forward passes into one paid dividends - it would be great to show what the 3 passes would cost if not batched.... And that, from a wall-clock time perspective, the ranking is: CA attenuation, Wen et al, CA-in-GUARD, Han et al and then Ren et al.}

% \begin{table}[h]
% \centering
% \caption{Average wall-clock runtime (in seconds) of each method across three experimental settings.}
% \label{tab:itmm-runtime}
% \footnotesize
% % \resizebox{\textwidth}{!}{%
% \resizebox{\columnwidth}{!}{%
% \begin{tabular}{l|ccc}
% \toprule
% % Method & SD v1.4 - verbatim mem. & SD v1.4 - template mem. & SD v2.0 - template mem. \\
% Method 
% & \shortstack{SD v1.4 \\ verbatim mem.} 
% & \shortstack{SD v1.4 \\ template mem.} 
% & \shortstack{SD v2.0 \\ template mem.} \\
% \midrule
% No mitigation & 543.677 & 1091.867 & 660.483 \\
% RTA          & 543.028 & 1072.585 & 660.944 \\
% Wen et al.   & 655.231 & 1170.091 & 836.057 \\
% Ren et al.   & 1046.507 & 2201.057 & 972.029 \\
% Han et al.   & 861.022 & 1709.124 & 1046.305 \\
% \midrule
% CA attenuation   & 572.986 & 1141.121 & 691.451 \\
% CA-in-GUARD      & 789.726 & 1567.801 & 932.459 \\
% \bottomrule
% \end{tabular}
% }
% \end{table}

\begin{table}[htb]
\centering
\caption{Average wall-clock runtime per example (in seconds) and GPU memory usage (in GiB) of each method across both model architectures.}
\label{tab:itmm-runtime}
\footnotesize
% \resizebox{\textwidth}{!}{%
\resizebox{\columnwidth}{!}{%
\begin{tabular}{l|cc|cc}
\toprule
\multirow{2}{*}{Method}
& \multicolumn{2}{c|}{SD v1.4}
& \multicolumn{2}{c}{SD v2.0} \\
& Time (s) & Memory (GiB)
& Time (s) & Memory (GiB) \\
\midrule
No mitigation  & 7.575  & 6.172  & 6.880  & 6.580 \\
RTA            & 7.563  & 6.172  & 6.884  & 6.590 \\
Wen et al.     & 9.210  & 12.697 & 8.709  & 13.102 \\
Ren et al.     & 15.302 & 6.213  & 10.125 & 6.695 \\
Han et al.     & 12.107 & 14.373 & 10.901 & 14.785 \\
\midrule
CA attenuation & 7.770  & 6.142  & 7.095  & 6.625 \\
CA-in-GUARD    & 10.963 & 6.543  & 9.713  & 6.735 \\
\bottomrule
\end{tabular}
}
\end{table}

% \kairan{
% \FloatBarrier
\subsection{Evaluation on SD v3.0}
\label{app:sdv3-evaluation}
Our main experiments follow prior work and evaluate memorization mitigation on SD v1.4 and SD v2.0. We additionally evaluate SD v3.0 to test whether the empirical gains of CA-in-GUARD extend to a newer and substantially different architecture. 
Unlike SD v1.4 and SD v2.0, which use U-Net-based denoising networks, SD v3.0 uses a transformer-based backbone. As a result, some prior methods and some parts of our mechanistic analysis do not transfer directly. In particular, \citet{ren2024unveiling} is not applicable in this setting and is therefore excluded.

We also note that the memorized prompt sets used in prior work were constructed for earlier Stable Diffusion models. Consequently, SD v3.0 exhibits substantially weaker memorization on this benchmark: only 39 of the 500 prompts satisfy $\mathrm{SSCD}>0.3$. We therefore evaluate SD v3.0 on this subset. 
% Because SD v3.0 differs architecturally from the models studied in the main paper, these results should be interpreted as an exploratory robustness check rather than a systematic study of memorization mechanisms in transformer-based diffusion models.

Table \ref{tab:sdv3-results} reports the results. CA-in-GUARD achieves the lowest SSCD among all applicable methods, while also obtaining the best FID and comparable CLIP. CA attenuation is also competitive: although its SSCD is slightly higher than \citet{han2025adjusting}, it achieves better CLIP and FID, indicating a favorable overall trade-off. 
These results suggest that the empirical benefits of our approach extend beyond the standard SD v1.4/v2.0 evaluation setting. 
However, because memorization may arise differently in transformer-based diffusion models, we leave a more systematic study of memorization mechanisms and mitigation strategies for SD v3.0-style architectures to future work.

\begin{table}[htb]
\centering
\caption{SD v3.0 evaluation results, extending Table \ref{tab:main_results_shorter}. For each method, we report the configuration that achieves the best SSCD.}
\label{tab:sdv3-results}
\footnotesize
\resizebox{\columnwidth}{!}{%
\begin{tabular}{l|ccc}
\toprule
Method & SSCD ($\downarrow$) & CLIP ($\uparrow$)  & FID ($\downarrow$) \\
\midrule
No mitigation  & $0.337{\pm}0.011$ & $0.213{\pm}0.021$ & $336.888$ \\
RTA            & $0.286{\pm}0.021$ & $0.182{\pm}0.023$ & $275.955$ \\
Wen et al.     & $0.303{\pm}0.031$ & $0.182{\pm}0.022$ & $373.299$ \\
Han et al.     & $0.200{\pm}0.037$ & $0.182{\pm}0.022$ & $219.990$ \\
\midrule
CA attenuation & $0.204{\pm}0.032$ & $0.185{\pm}0.027$ & $215.708$ \\
CA-in-GUARD    & $0.168{\pm}0.017$ & $0.181{\pm}0.020$ & $212.728$ \\
\bottomrule
\end{tabular}}
\end{table}
% }

% \kairan{
\subsection{Qualitative Examples}
\label{app:qualitative-examples}

We provide additional qualitative examples across model architectures and memorization settings, as shown in Figure \ref{fig:qualitative}, to complement the quantitative results in Figure \ref{fig:qualitative-main} in the main paper. These examples compare generations from the no-mitigation baseline, prior mitigation methods, CA attenuation, and CA-in-GUARD against the corresponding training images.

Across examples, the no-mitigation baseline often preserves substantial visual overlap with the training image, especially in layout, object identity, and distinctive local details.
Prior mitigation methods reduce this similarity to varying degrees, reflecting different trade-offs between memorization mitigation and generation quality. 
CA-in-GUARD shows the most consistent visual trade-off: it produces generations that are visually distinct from the corresponding training images while preserving prompt alignment and overall image quality. 
This trend is visible in both verbatim and template memorization settings. 
Notably, even under template memorization, which is harder to mitigate quantitatively, CA-in-GUARD still substantially moves the generated image away from the memorized training example while maintaining good generation quality.

These qualitative results, together with the SSCD, CLIP, FID, and DINO results reported above, provide additional evidence that CA-in-GUARD mitigates memorization without relying on severe degradation of generation quality.
\begin{figure*}[htb]
\centering
\begin{subfigure}[htb]{\textwidth}
 \centering
\includegraphics[scale=0.25]{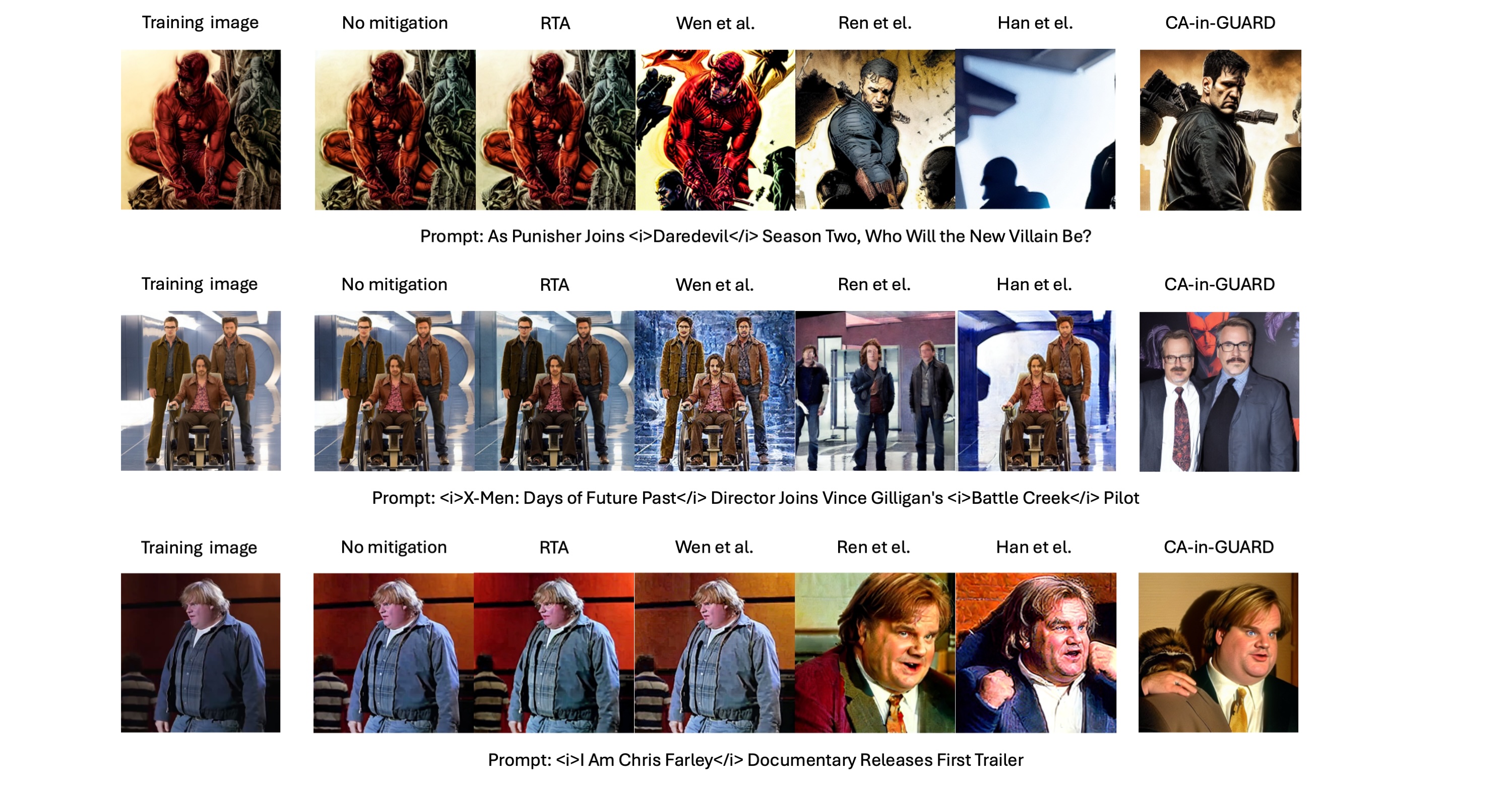}
 \caption{SD v1.4 - verbatim memorization}
 % \label{fig:itmm-sscd}
\end{subfigure}
%  \hfill
\begin{subfigure}[htb]{\textwidth}
\centering
\includegraphics[scale=0.25]{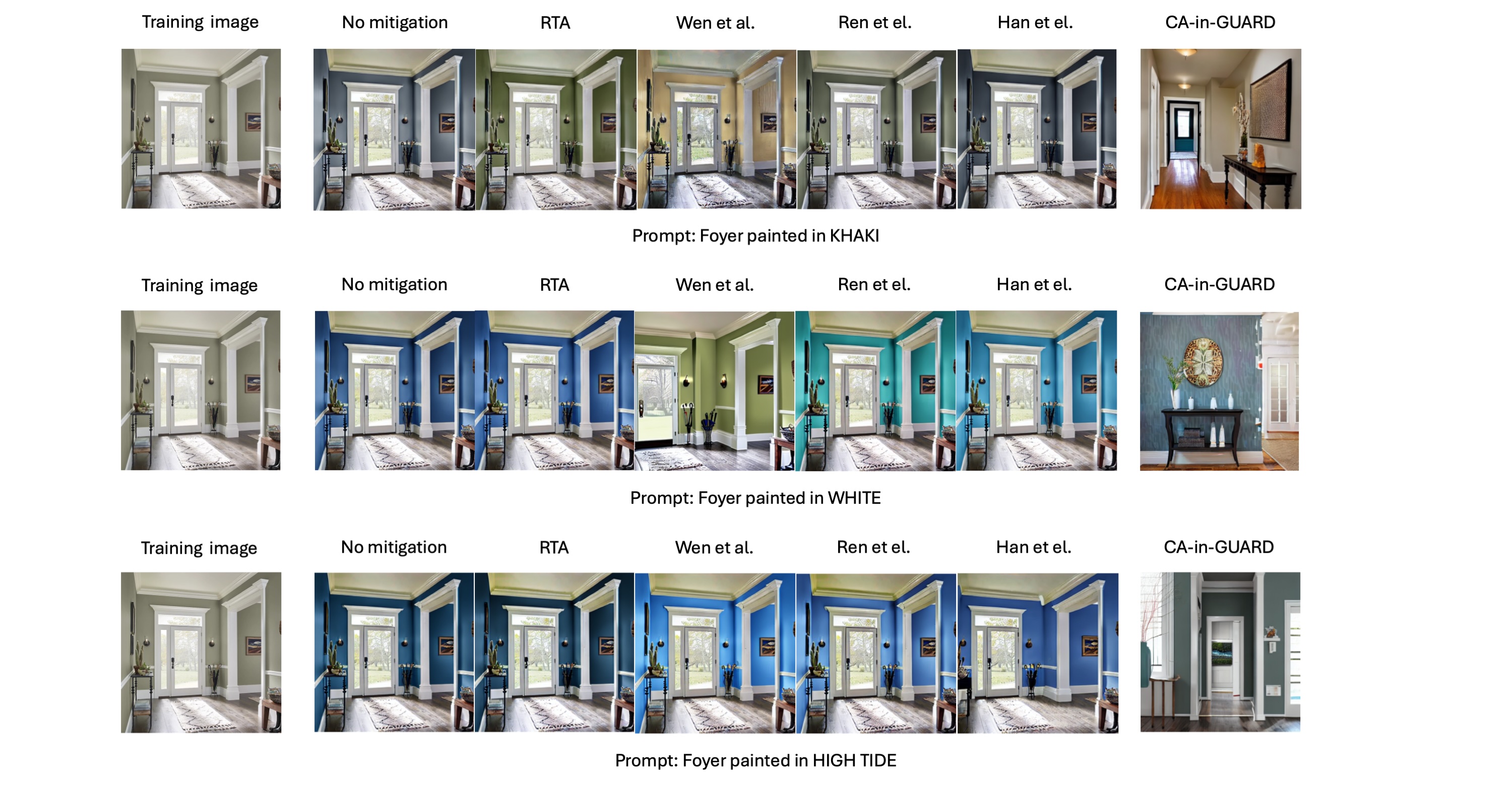}
\caption{SD v1.4 - template memorization}
% \label{fig:itmm-clip}
\end{subfigure}
%  \hfill
\begin{subfigure}[htb]{\textwidth}
\centering
\includegraphics[scale=0.25]{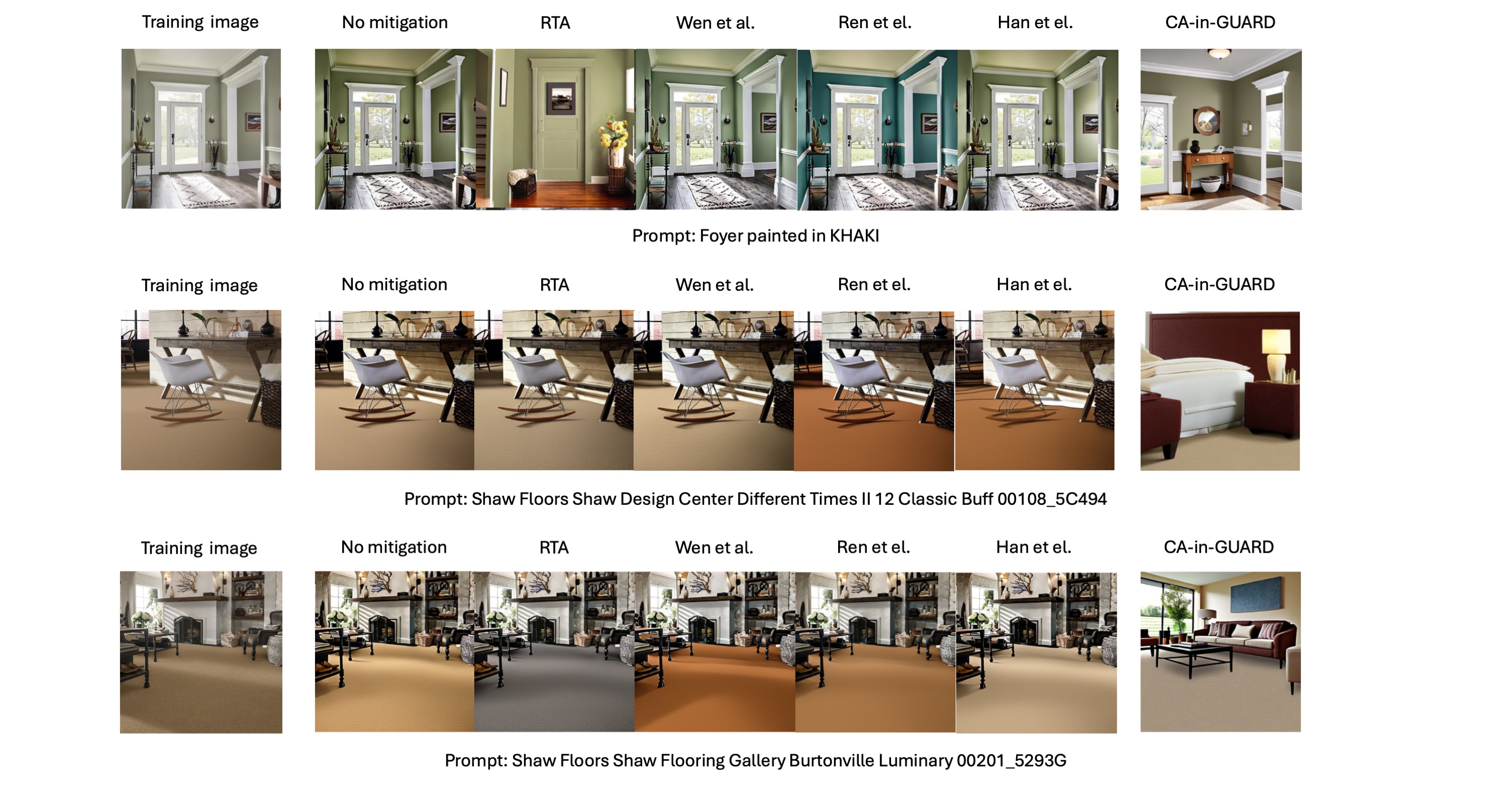}
\caption{SD v2.0 - template memorization}
% \label{fig:itmm-clip}
\end{subfigure}
%  \hfill
\caption{Additional qualitative examples across model architectures and memorization settings. We compare the training image with generations from no mitigation, prior mitigation methods, CA attenuation, and CA-in-GUARD. CA-in-GUARD consistently moves generations away from memorized training images while preserving prompt-relevant visual content.}
\label{fig:qualitative}
\end{figure*}
% }

\end{document}